\documentclass{article} % For LaTeX2e
\usepackage[OT1]{fontenc}
\usepackage{iclr2027_conference,times}

\usepackage{amsmath,amsfonts,bm}

\def\eqref#1{equation~\ref{#1}}
\def\1{\bm{1}}

\DeclareMathAlphabet{\mathsfit}{\encodingdefault}{\sfdefault}{m}{sl}
\SetMathAlphabet{\mathsfit}{bold}{\encodingdefault}{\sfdefault}{bx}{n}

\usepackage{hyperref}
\hypersetup{hidelinks}
\usepackage{url}
\usepackage{amssymb}
\usepackage{booktabs}
\usepackage{multirow}
\usepackage{tabularx}
\usepackage{array}
\usepackage{graphicx}
\usepackage{float}
\usepackage{caption}
\usepackage{wrapfig}
\usepackage{tikz}
\usetikzlibrary{arrows.meta,fit,positioning}

\title{WorldProbe: When Video World Models Fall Apart under Long-Horizon Spatial Exploration}

\author{
\textbf{Minqi Wang}$^{1,2}$, \textbf{Jie Peng}$^{3}$, \textbf{Hejun Dong}$^{4}$, \textbf{Ruchang Yao}$^{1}$, \textbf{Puyi Wang}$^{1}$, \textbf{Siyuan Huang}$^{1}$, \\
\textbf{Jihai Zhang}$^{1}$, \textbf{Yafu Li}$^{1}$, \textbf{Qi Dou}$^{1}$, \textbf{Xiaoye Qu}$^{5}$\thanks{Corresponding authors.}, \textbf{Yu Cheng}$^{6}$\footnotemark[1] \vspace{1mm} \\
$^{1}$The Chinese University of Hong Kong \quad $^{2}$Peking University \\
$^{3}$University of Science and Technology of China \\
$^{4}$Department of Computer Science and Engineering, CUHK \\
$^{5}$Shanghai Artificial Intelligence Laboratory \quad $^{6}$Nanyang Technological University \\
{\small\texttt{mikeywang512@gmail.com, quxiaoye@pjlab.org.cn, yu.cheng@ntu.edu.sg}}
}

\iclrfinalcopy
\begin{document}

\maketitle
\lhead{Preprint}
\suppressfloats[t]

\begin{abstract}
Video world models generate future visual observations from camera or action inputs, with the goal of synthesizing persistent 3D worlds that can be explored across time and space. Existing benchmarks typically evaluate short interactions or rely on camera trajectories fixed before generation. The performance of video world models under long-horizon exploration with extensive camera motion remains under-explored. We introduce \textbf{WorldProbe}, a diagnostic benchmark that uses a goal-conditioned multimodal agent to adapt camera actions online during generation. WorldProbe comprises 164 test cases across four probing families. Across 10 state-of-the-art models, we generate more than 1,600 rollouts totaling approximately 800 minutes, with horizons up to 180 seconds. Our evaluation across four dimensions reveals both the degradation of local measures and the emergence of long-range inconsistencies. We find that recent models sustain visual and content quality for longer, while long-range memory and evolution remain weak. These results expose a gap between locally stable generation and persistent world consistency that is largely missed by existing benchmarks. Code and benchmark resources are available at \url{https://github.com/Untitled-W/WorldProbe}.
\end{abstract}

\begin{figure}[!ht]
\centering
\includegraphics[width=\linewidth]{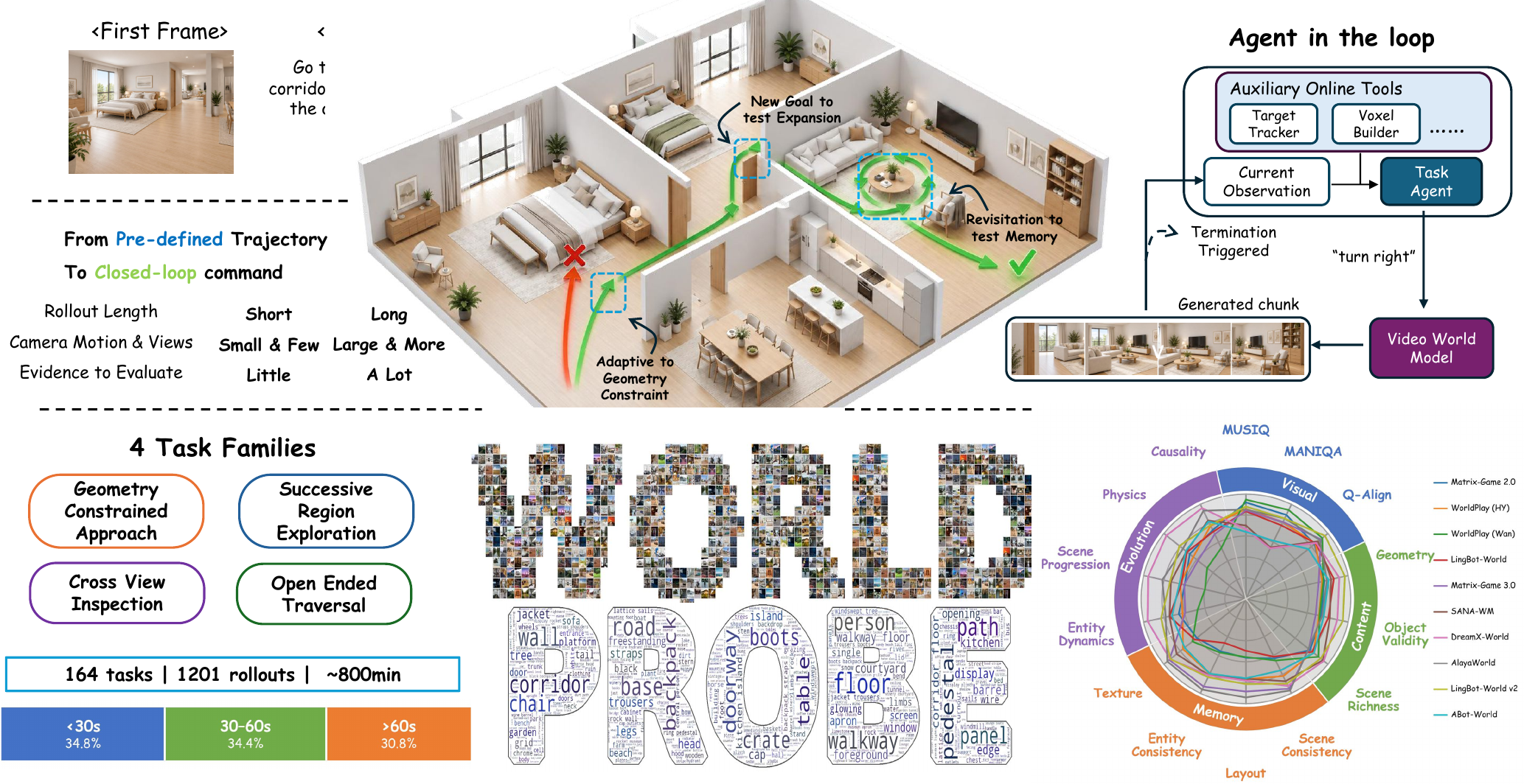}
\caption{\textbf{WorldProbe overview.} A goal-conditioned agent adaptively probes an evolving video world through closed-loop camera control, enabling long-horizon evaluation across four task families and four evaluation dimensions.}
\label{fig:overview}
\end{figure}

\section{Introduction}
\label{sec:introduction}

Video generation is moving from fixed-sequence synthesis toward interactive world simulation~\citep{openai2024sora,bruce2024genie,deepmind2024genie2,deepmind2025genie3}. Instead of passively viewing a predetermined video, users can steer the camera to dynamically shape subsequent rollouts~\citep{zhang2025matrixgame}. Beyond visual quality improvements, recent advances in autoregressive architectures and long-context modeling enable such interactions to unfold over extended horizons and flexible spatial trajectories~\citep{wang2026matrixgame3,gao2026lingbotworld,wu2026infiniteworld,sun2025worldplay}.

\begin{table}[t]
\vspace{-8pt}
\setlength{\abovecaptionskip}{1pt}
\setlength{\belowcaptionskip}{0pt}
\centering
\caption{\textbf{Comparison with existing video world-model benchmarks.} \textbf{Int.} interactive control. \textbf{Cplx.} probing complexity: T = text prompt; CC = case-by-case design; TC = template composition; AA = agent action. $^{*}$ inferred from the dataset.}
\label{tab:benchmark_comparison}
\scriptsize
\setlength{\tabcolsep}{1.5pt}
\renewcommand{\arraystretch}{0.88}
\begin{tabular*}{\linewidth}{@{\extracolsep{\fill}}lcccc@{}}
\toprule
\multicolumn{1}{c}{Benchmark} & Int. & Cplx. & Len. & $N$ \\
\midrule
\mbox{WorldModelBench~\citep{li2025worldmodelbench}} & $\times$ & Low (T) & $\sim$10\,s$^{*}$ & 350 \\
\mbox{WorldScore~\citep{duan2025worldscore}} & $\times$ & Mid (CC) & $\leq$10\,s$^{*}$ & 3000 \\
\mbox{WorldMark~\citep{worldmark2026}} & $\times$ & Mid (TC) & 60\,s & 500 \\
\mbox{WBench~\citep{wbench2026}} & $\times$ & Mid (CC) & $\leq$9 tr. & 289 \\
\mbox{LoopNav~\citep{lian2025loopnav}} & $\times$ & Mid (TC) & $\sim$1200\,fr. & 144 \\
\mbox{Omni-WorldBench~\citep{wu2026omniworldbench}} & $\times$ & Mid (CC) & $\sim$5.7\,s$^{*}$ & 1068 \\
\mbox{iWorld-Bench~\citep{fang2026iworldbench}} & $\times$ & Mid (TC) & 7.6\,s & 4900 \\
\midrule
\mbox{\textbf{WorldProbe}} & $\checkmark$ & High (AA) & 180\,s & 164 \\
\bottomrule
\end{tabular*}
\vspace{-4pt}
\end{table}
\noindent Yet a world model should not merely render plausible frame sequences. It should instead sustain a persistent, evolving world. The scene should remain consistent upon revisitation, expand coherently into unseen regions, and follow camera commands without semantic drift~\citep{sun2025worldplay,ren2025gen3c,hunyuanworld2025,yang2025matrix3d}. However, current benchmarks provide limited evidence in this regime~\citep{li2025worldmodelbench,duan2025worldscore,worldmark2026,wbench2026,lian2025loopnav,wu2026omniworldbench,fang2026iworldbench}. This stems from three limitations. First, models are usually evaluated on short rollouts with limited camera motion. Failures in memory, consistency, and scene evolution often emerge only under long-horizon exploration with substantial camera motion. Second, current benchmarks rely on camera trajectories fixed before generation. Such trajectories cannot adapt to newly revealed geometry or redirect exploration toward newly selected targets, limiting the complexity of probing tasks. Third, existing metrics primarily assess local cues and are typically computed over neighboring frames or views with geometric priors or learned scorers for control and image quality~\citep{huang2023vbench,huang2024vbenchpp,zheng2025vbench2,qin2024worldsimbench,wu2026omniworldbench}. Existing memory tests extend the temporal gap, but still rely on pairwise matching at identical camera poses~\citep{worldmark2026,lian2025loopnav}. As a result, these metrics do not naturally capture the long-range failures that emerge over extended rollouts. Table~\ref{tab:benchmark_comparison} summarizes how existing benchmarks compare along interaction, trajectory complexity, horizon, and scale.

Building on these insights, we introduce \textbf{WorldProbe}, a benchmark for closed-loop agentic probing of video world models. Each probe starts from an initial frame and a spatial objective designed to elicit diagnostic camera motion. A goal-conditioned multimodal agent adaptively controls the camera based on ongoing generation to fulfill the objective. Because current multimodal models remain limited in spatial reasoning, we introduce auxiliary perception modules to stabilize long-range goal pursuit. In extended rollouts, the probing loop remains stable for up to 5 minutes and ends only when the generated video itself collapses.

\begin{figure*}[t]
\centering
\includegraphics[width=\textwidth,trim=47bp 37bp 49bp 37bp,clip]{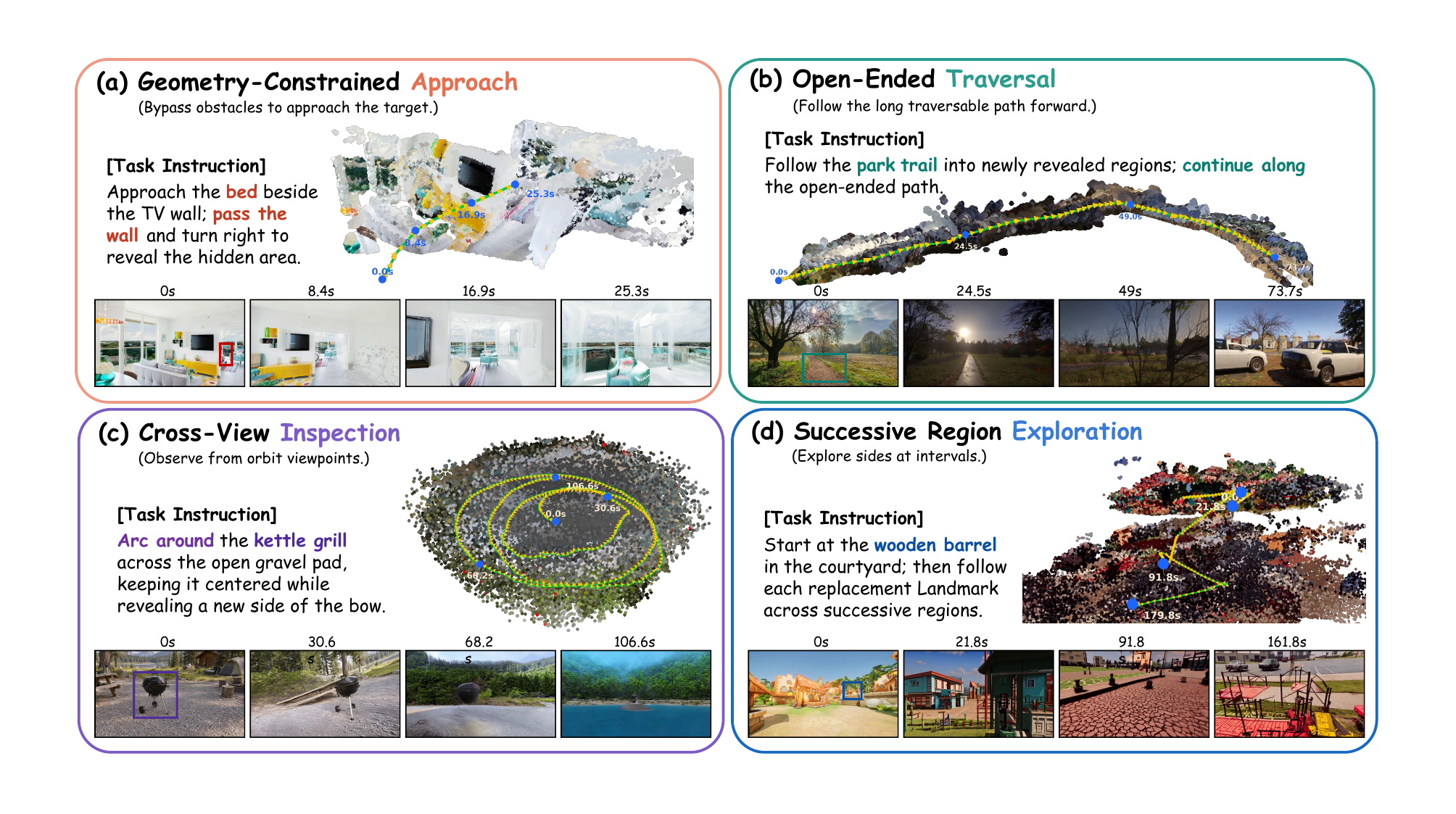}
\setlength{\abovecaptionskip}{4pt}
\caption{\textbf{Four scene-adaptive probing tasks.} Each task induces a distinct spatial exploration pattern: approaching a constrained target, traversing an open path, inspecting a scene across viewpoints, or successively exploring new regions.}
\label{fig:tasks}
\end{figure*}

We construct 164 test cases across four spatial probing families: geometry-constrained approach, open-ended traversal, cross-view inspection, and successive region exploration. We characterize long-horizon failure from two complementary perspectives: degradation of local measures and exposure of long-range inconsistencies. The former tracks how local quality degrades over time, revealing long-term stability and drift. The latter captures long-range contradictions and semantic errors that local measures miss. We organize the evaluation along four dimensions: visual quality, content quality, long-range memory, and evolution fidelity.

We evaluate 10 state-of-the-art video world models over more than 1,600 rollouts, totaling approximately 800 minutes with horizons up to 180 seconds. We observe two consistent failure patterns. (1) Visual and content quality significantly degrade as the rollout proceeds. Newer models often degrade more slowly, indicating progress in long-horizon generation. (2) All models also exhibit substantial long-range inconsistencies. In memory tests, simple leave-and-return trajectories may preserve coarse geometry while losing texture, whereas revisiting from a new direction often produces an entirely different scene. During evolution, objects may split, collapse, vanish, or duplicate. Objects may also move inconsistently with the camera. Scenes may repeat structures that should not recur or drift into unrelated content. These failures are largely missed by existing short-horizon benchmarks.

Our contributions are summarized as follows:
\begin{itemize}
    \setlength{\parskip}{0pt}
    \item \textbf{Long-horizon evaluation protocol.}
    We introduce a closed-loop agentic probing protocol that extends video world model evaluation to long-horizon spatiotemporal interaction.
    \item \textbf{Goal-conditioned agent design.}
    We develop a goal-conditioned multimodal agent that adaptively controls camera motion during generation, together with auxiliary modules that stabilize long-range goal pursuit.
    \item \textbf{Benchmark and metrics.}
    We construct 164 test cases across four spatial probing families and evaluate long-horizon behavior from two complementary perspectives across four dimensions: visual quality, content quality, long-range memory, and evolution fidelity.
    \item \textbf{Large-scale empirical study.}
    We evaluate 10 state-of-the-art video world models and uncover systematic long-horizon failures that are largely missed by existing benchmarks.
\end{itemize}

\section{Related Work}
\label{sec:related_work}

\subsection{Interactive Video World Models}

Most video generation models synthesize a complete sequence from a fixed prompt and do not accept camera actions during generation~\citep{openai2024sora,wanteam2025wan,kong2024hunyuanvideo,yang2024cogvideox}. Video world models instead condition future observations on camera or action inputs, allowing the generated sequence to respond to interaction~\citep{bruce2024genie,deepmind2024genie2,deepmind2025genie3}. Subsequent models expanded to broader scenes and richer control signals~\citep{zhang2025matrixgame,li2025hunyuan_gamecraft,tang2025hunyuan_gamecraft2,mao2025yume,mao2025yume15}. History-conditioned and memory-based models retain or retrieve earlier observations to support longer rollouts~\citep{wang2026matrixgame3,gao2026lingbotworld,wu2026infiniteworld,sun2025worldplay,alayaworld2026,lingbotworld22026,matrixgame352026,wonder2026}. Geometry-aware models use camera cues, explicit 3D structure, or spatial memory to preserve scenes across viewpoint changes and revisits~\citep{ren2025gen3c,hunyuanworld2025,yang2025matrix3d,gimworld2026,mirage2026,worlddirector2026,worldinworld2026}. These models enable long-range interaction, while their ability to maintain a coherent world at these scales remains an open evaluation question.

\subsection{Video World Model Evaluation}

General video benchmarks measure perceptual quality and conditional faithfulness~\citep{huang2023vbench,huang2024vbenchpp,zheng2025vbench2,sun2024t2vcompbench}. World model benchmarks further evaluate physical plausibility, world dynamics, and responses to camera or action inputs~\citep{li2025worldmodelbench,qin2024worldsimbench,duan2025worldscore,wu2026omniworldbench,fang2026iworldbench}. Table~\ref{tab:benchmark_comparison} consolidates how these benchmarks differ in interaction, probing complexity, horizon, and scale. WorldMark uses predefined trajectories of up to 60 seconds, while WBench averages 3.7 interaction turns and has only 12\% of cases beyond four turns~\citep{worldmark2026,wbench2026}. LoopNav reaches 50 Minecraft blocks but follows prescribed loops and evaluates only their return segments~\citep{lian2025loopnav}. These settings provide useful tests for selected trajectories, yet offer limited adaptation to generated geometry and spatial goals. In contrast, WorldProbe evaluates up to 180 seconds of closed-loop probing, with trajectories adapted online from generated observations under a task-specific objective.

\section{WorldProbe}
\label{sec:method}

WorldProbe first generates task-conditioned rollouts through closed-loop agent interaction and then evaluates them with the proposed metrics, as illustrated in Fig.~\ref{fig:overview}.

\subsection{Probing Task Design}
\label{sec:task_design}

Different spatial interactions expose different failure modes. WorldProbe therefore defines four scene-adaptive probing tasks, each specifying a spatial objective while allowing the agent to adapt its actions to the generated scene. Representative examples are shown in Fig.~\ref{fig:tasks}.

\begin{itemize}
    \setlength{\parskip}{0pt}
    \item \textbf{Geometry Constrained Approach.}
    The agent approaches a partially visible target while avoiding surrounding structures. This task tests whether the model completes the target and keeps newly revealed geometry consistent with earlier observations.
    \item \textbf{Open-Ended Traversal.}
    The agent follows an unbounded scene structure, such as a road, corridor, or pathway, without a predefined endpoint. This task tests long-range continuity and the preservation of scene layout along an extended path.
    \item \textbf{Cross-View Inspection.}
    The agent observes the same local scene from distinct viewpoints. This task tests multi-view consistency and the generation of content revealed from new views.
    \item \textbf{Successive Region Exploration.}
    The agent successively redirects toward new peripheral regions. This task tests whether the model can combine locally revealed regions into a coherent environment while preserving previously generated content.
\end{itemize}

We construct task instances by matching the spatial concept of each task with suitable initial frames from public datasets \citep{ye2026mind,duan2025worldscore,worldmark2026}. We then manually write a task prompt for each selected frame, specifying the spatial objective and the intended camera behavior. The resulting cases span diverse indoor and outdoor scenes.

\subsection{Agent Design}
\label{sec:agent_design}

WorldProbe uses a goal-conditioned multimodal agent for camera control. At each decision step, the agent receives the task specification, a short window of recent frames and auxiliary spatial states. The agent predicts one action from a discrete action space. The selected action is converted to the model-specific control format and applied to the next generation chunk. The resulting frames are added to the temporal context for the next decision.

\noindent Long-horizon control requires the task target to remain visually grounded throughout the rollout. A text-only target description is insufficient once the scene changes substantially, likely due to the limited spatial reasoning capability of current multimodal models. We therefore maintain an online segmentation tracker that updates the target location for each generated chunk. We use SAM3.1 for its memory mechanism and text-conditioned tracking, which help preserve target identity across changing observations. Appearance-based trackers are less robust under the large appearance and geometric changes induced by long rollouts. A geometric gate filters masks with implausibly large frame-to-frame displacement.

\begin{wraptable}[18]{r}{0.60\textwidth}
\vspace{-8pt}
\noindent\begin{minipage}{0.60\textwidth}
\setlength{\abovecaptionskip}{1pt}
\setlength{\belowcaptionskip}{0pt}
\captionsetup{width=\linewidth}
\centering
\caption{\textbf{Video world model inference configuration.} Per-chunk latency and rollout statistics for evaluated models.}
\label{tab:runtime}
\small
\renewcommand{\arraystretch}{1.05}
\begin{tabular}{@{}l@{\hspace{11.5pt}}c@{\hspace{11.5pt}}c@{\hspace{11.5pt}}c@{\hspace{11.5pt}}c@{\hspace{11.5pt}}c@{}}
\toprule
\multicolumn{1}{c}{\multirow{2}{*}{Model}} & \makebox[0pt]{Latency} & \makebox[0pt]{Chunk$^{*}$} & \multirow{2}{*}{\begin{tabular}[c]{@{}c@{}}Avg.\\[-1.5pt]Len\end{tabular}} & \multirow{2}{*}{\begin{tabular}[c]{@{}c@{}}Pass\\[-1.5pt]Rate\end{tabular}} & \multirow{2}{*}{$>$60s} \\
& \makebox[0pt]{\scalebox{0.86}{\scriptsize (per chunk)}} & \makebox[0pt]{\scalebox{0.86}{\scriptsize (frames/FPS)}} & & & \\
\midrule
Matrix-Game 2.0 & 1.469s & 12/12 & 54.4s & 73.2\% & 26.8\% \\
\mbox{WorldPlay (HY)} & 14.651s & 16/24 & 59.6s & 81.1\% & 24.2\% \\
WorldPlay (Wan) & 3.418s & 16/16 & 42.4s & 57.9\% & 9.8\% \\
LingBot-World & 6.681s & 12/16 & 39.5s & 48.8\% & 12.4\% \\
Matrix-Game 3.0 & 8.178s & 40/17 & 49.5s & 62.2\% & 17.6\% \\
SANA-WM & 7.042s & 24/16 & 40.8s & 65.9\% & 8.5\% \\
DreamX-World & 3.514s & 12/16 & 42.9s & 56.1\% & 12.4\% \\
AlayaWorld & 8.224s & 32/24 & 56.9s & 89.6\% & 26.8\% \\
LingBot-World v2 & 8.362s & 16/16 & 52.7s & 69.5\% & 26.1\% \\
ABot-World & 3.233s & 12/12 & 51.3s & 65.9\% & 26.1\% \\
\bottomrule
\end{tabular}
\vspace{2pt}

\raggedright\scriptsize\textit{* Chunk reports the frame count per chunk and the video generation FPS. Frame count denotes subsequent chunks only; the first chunk is excluded.}
\end{minipage}
\vspace{-4pt}
\end{wraptable}
\noindent Target grounding gives the agent a target direction but does not determine the distance to nearby obstacles. The agent can still underestimate clearance and continue forward after reaching a required turning point. Inspired by MG3.5~\citep{matrixgame352026}, we add an explicit 3D geometry reconstruction module to avoid severe collisions. For each chunk, Depth-Anything3 estimates depth from which we construct a coarse voxelized geometry. We use it to estimate free-space distance along the current forward direction. If this distance is no greater than the forward displacement in the last chunk, the controller overrides the predicted action and turns toward the current target direction. We restrict reconstruction to chunk-level geometry, using it only for collision avoidance while leaving global navigation to the agent. This separation avoids costly long-range 3D reconstruction and the ambiguity that arises when accumulated global geometry becomes inconsistent with the current generated observations.

\begin{wraptable}[9]{r}{0.35\textwidth}
\vspace{-8pt}
\setlength{\abovecaptionskip}{1pt}
\setlength{\belowcaptionskip}{0pt}
\captionsetup{width=\linewidth}
\centering
\caption{\textbf{Additional process\-ing latency.}}
\label{tab:ad-processing}
\small
\setlength{\tabcolsep}{2pt}
\renewcommand{\arraystretch}{1.0}
\begin{tabular*}{\linewidth}{@{\extracolsep{\fill}}>{\centering\arraybackslash}m{0.58\linewidth}>{\centering\arraybackslash}m{0.28\linewidth}@{}}
\toprule
Module & Latency \\
\multicolumn{1}{>{\centering\arraybackslash}m{0.58\linewidth}}{} & \scriptsize (per chunk) \\
\midrule
\multicolumn{1}{>{\centering\arraybackslash}m{0.58\linewidth}}{Qwen reasoning} & 3.025s \\
\multicolumn{1}{>{\centering\arraybackslash}m{0.58\linewidth}}{SAM3.1 tracking} & 7.609s \\
\multicolumn{1}{>{\centering\arraybackslash}m{0.58\linewidth}}{DA3+ Voxel} & 0.5143s \\
\bottomrule
\end{tabular*}
\vspace{-4pt}
\end{wraptable}
\noindent Before action prediction, the agent performs a compact structured reasoning step over target direction and intended motion rather than producing free-form chain-of-thought. This reduces reasoning latency by over 90\% without measurable loss in task success. Table~\ref{tab:ad-processing} shows that the auxiliary modules add modest overhead.

The agent may explicitly terminate a stage once it determines that the spatial objective has been completed. We also terminate unrecoverable rollouts based on persistent tracking failure or oscillatory control. After generation, we filter unsuccessful or invalid rollouts using minimum-duration criteria and coarse VLM screening for severe visual or trajectory anomalies.

\subsection{Metrics Design}
\label{sec:metrics}

Our metrics characterize long-horizon failure from two complementary perspectives: degradation of local measures and exposure of long-range inconsistencies. Local measures track visual and content quality over individual frames, while long-range metrics assess memory and evolution fidelity over extended temporal windows.

\noindent\textbf{Local visual and content quality.}
We evaluate visual quality with three no-reference image-quality assessment (IQA) models: MUSIQ, MANIQA, and Q-Align. Each score is normalized from its theoretical range to 0--1 before averaging as the overall visual quality score. For content quality, we leverage a Vision-Language Model (VLM) to rate geometric validity of visible object relations, object validity, and scene richness on a 0--10 integer scale. Each rating is computed independently from a single frame, without earlier-frame context or task-specific target annotations. Frames are sampled approximately uniformly over the rollout, with the sampling density adjusted to video duration.

\noindent\textbf{Long-term memory and evolution.}
For memory evaluation, we use ViPE~\citep{huang2025vipe} to estimate the camera pose sequence from the generated frames. We then align the full camera-pose sequence to a 1-fps candidate track and use farthest-point sampling to select up to five spatially separated viewpoints with a 3-m minimum distance between anchors. At each anchor, we select up to eight temporally ordered frames within 30 degrees and 3 m, with estimated FoV overlap above 0.30 and at least 8 s between any pair. We leverage a VLM to score in 4 dimensions: scene consistency, layout consistency, entity consistency, and texture consistency from 0 to 10 for each supported viewpoint. Scores are averaged across supported viewpoints and then across rollouts.

Compared with prior work~\citep{sun2025worldplay,worldmark2026,lian2025loopnav,zhang2026mbench}, we make four design choices to more faithfully evaluate spatial memory. First, unlike prior methods built around predefined loop trajectories, our evaluation applies to arbitrary trajectories. Second, we anchor frame selection using visually reconstructed rather than commanded camera poses, since accumulated control-following errors would otherwise conflate memory with action fidelity. Third, we jointly compare multiple frames over an extended temporal window, shifting the evaluation from frame matching toward spatial memory. Finally, we compare nearby but non-identical viewpoints, making the evaluation robust to moderate viewpoint changes.

For evolution fidelity, we evaluate four dimensions: entity evolution, scene evolution, physical plausibility, and action--outcome causality. We use the same temporal sampling strategy as the local metrics. The sampled frames and a summary of the control sequence are jointly provided to a VLM, which produces rollout-level scores for the four dimensions.

\section{Experiments}
\label{sec:experiments}

\subsection{Experimental Setup}
\label{sec:setup}

\begin{wraptable}[19]{r}{0.58\textwidth}
\vspace{-10pt}
\setlength{\abovecaptionskip}{1pt}
\setlength{\belowcaptionskip}{0pt}
\captionsetup{width=\linewidth}
\centering
\caption{\textbf{Local quality and temporal degradation across models.} Mean measures average rollout-level quality, while Slope measures its change over time; more negative slopes indicate faster degradation. Visual scores are normalized to 0--1, and content scores retain their 0--10 scale. Visual slopes are reported in $10^{-3}$/s; content slopes are in points/s.}
\label{tab:local-metrics}

\footnotesize

\setlength{\tabcolsep}{3pt}

\renewcommand{\arraystretch}{0.88}

\begin{tabular*}{\linewidth}{@{\extracolsep{\fill}}lcccc@{}}

\toprule

\multicolumn{1}{c}{\multirow{2}{*}{Model}} & \multicolumn{2}{c}{Visual} & \multicolumn{2}{c}{Content} \\

\cmidrule(lr){2-3}\cmidrule(lr){4-5}

& Mean & Slope & Mean & Slope \\

\midrule

\mbox{Matrix-Game 2.0} & 0.565 & $-0.77$ & 6.09 & $-0.061$ \\

\mbox{WorldPlay (HY)} & 0.542 & $-3.30$ & 6.08 & $-0.072$ \\

\mbox{WorldPlay (Wan)} & \textbf{0.603} & $-2.91$ & 6.29 & $-0.092$ \\

\mbox{LingBot-World} & 0.532 & $-2.27$ & 6.61 & $-0.026$ \\

\mbox{Matrix-Game 3.0} & 0.556 & \textbf{$+0.76$} & 5.76 & $-0.051$ \\

\mbox{SANA-WM} & 0.474 & $-3.32$ & 6.91 & $-0.054$ \\

\mbox{DreamX-World} & 0.458 & $-3.11$ & 7.36 & $-0.018$ \\

\mbox{AlayaWorld} & 0.560 & $-2.61$ & 6.14 & $-0.093$ \\

\mbox{LingBot-World v2} & 0.583 & $-1.02$ & \textbf{7.55} & \textbf{$-0.015$} \\

\mbox{ABot-World} & 0.439 & $-5.48$ & 6.39 & $-0.065$ \\

\bottomrule

\end{tabular*}

\vspace{-8pt}
\end{wraptable}
We evaluate 10 publicly available video world models~\citep{he2025matrixgame2,sun2025worldplay,gao2026lingbotworld,wang2026matrixgame3,zhu2026sanawm,dreamxworld2026,alayaworld2026,lingbotworld22026,jiang2026abotworld}, with their inference configurations summarized in Table~\ref{tab:runtime}. We use Qwen3.6-27B~\citep{qwen2026qwen36} as the VLM backbone for the task agent, Segment-Anything-3.1~\citep{carion2025sam3} for target tracking, and Depth-Anything-3~\citep{lin2025depthanything3} for local geometry reconstruction. We set task-specific rollout duration ranges of 10--60,s, 20--120,s, 20--90,s, and 30--180,s for the four task families, respectively. Rollouts below the minimum duration are retried up to three times, and the longest attempt is retained. Evaluating all models took approximately four days on 24 A100 GPUs.

\WFclear
We use a client--server inference architecture to support concurrent long-horizon rollouts. The VLM and world models remain resident as separate services, while the task agent coordinates closed-loop interaction through lightweight HTTP communication. We allocate 2--3 GPUs to the VLM service and the remaining GPUs to world-model workers.

We made two main implementation efforts to support long-horizon closed-loop evaluation. First, although these models conceptually support online control, most public implementations do not expose an interface for receiving new actions between generated chunks. We therefore refactor each inference pipeline into step-wise generation so that a new action can be injected after every chunk. Second, implementations tuned for short rollouts often become unstable beyond 30--60 seconds due to memory growth and incorrect cross-chunk state handling. We therefore revise memory and state handling to support stable long-horizon generation. Black-box comparisons under each repository's native setting confirm that these changes preserve the original model behavior.

\subsection{Local Measure Degradation}

\begin{figure}[!t]
\centering
\includegraphics[width=\linewidth]{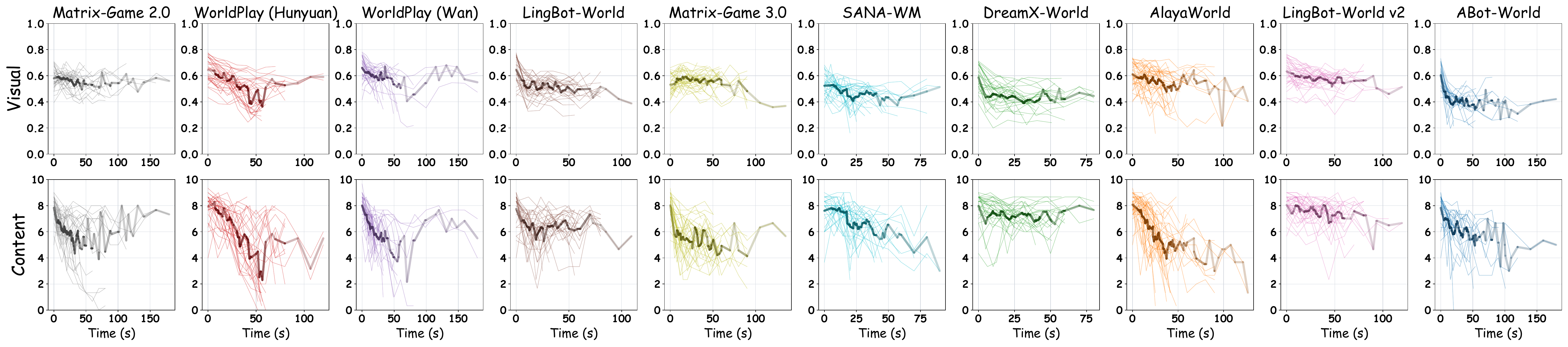}
\caption{\textbf{Local quality degrades over long-horizon interaction.} Per-rollout visual and content trajectories show stronger collapse in earlier models, while more recent models generally maintain a higher quality floor. Models are ordered by release time from left to right.}
\label{fig:local-degradation}
\end{figure}

We first examine how locally measurable quality changes as interaction extends. Fig.~\ref{fig:local-degradation} shows a clear temporal degradation across video world models, but with markedly different failure patterns. Models are ordered by release time from left to right. Earlier models frequently exhibit abrupt collapse: many rollouts remain plausible for an initial period but deteriorate rapidly beyond a certain horizon, with visual and content scores approaching zero. More recent models show substantially more stable trajectories. Although their quality still declines over time, they tend to maintain a non-trivial quality floor rather than collapsing completely. This suggests that recent progress has not eliminated long-horizon degradation, but has extended the effective generation horizon over which usable content can be sustained.

\begin{wrapfigure}[23]{r}{0.70\textwidth}
\vspace{-8pt}
\setlength{\abovecaptionskip}{1pt}
\setlength{\belowcaptionskip}{0pt}
\captionsetup{width=\linewidth}
\centering
\includegraphics[width=\linewidth]{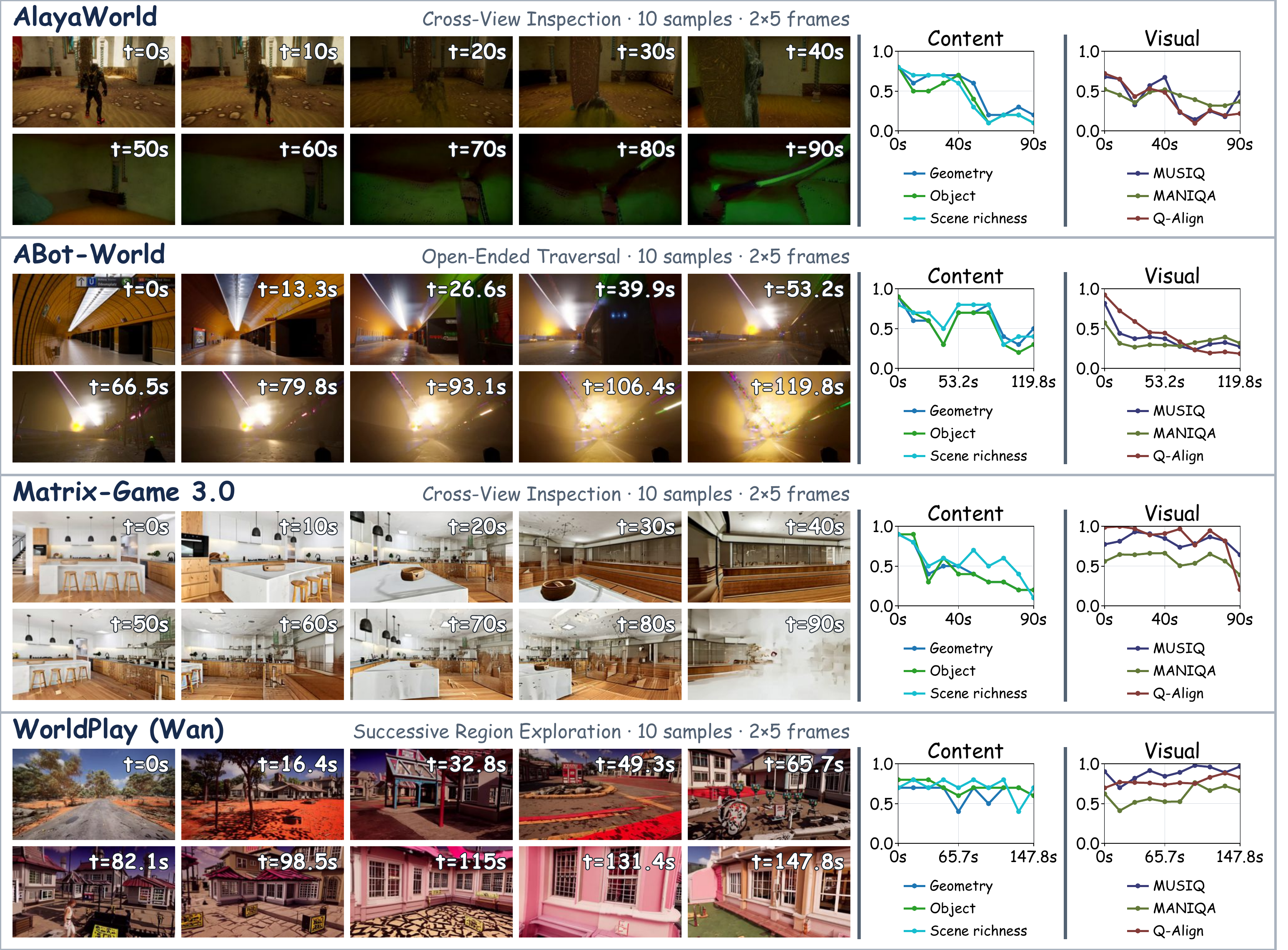}
\caption{\textbf{Representative local-quality behaviors.} Metric degradation aligns with visible quality loss, while some long rollouts sustain rich content without evident collapse.}
\label{fig:local-cases}
\vspace{-4pt}
\end{wrapfigure}
Table~\ref{tab:local-metrics} quantifies these trends through rollout-level means and temporal slopes. The degradation is consistently stronger for content quality than for visual quality. Across long rollouts, models often continue to produce perceptually plausible frames even after geometric validity, object integrity, and scene richness have deteriorated substantially. This gap suggests that maintaining visually acceptable rendering is easier than preserving semantically rich and structurally valid content over extended generation. Mean quality and degradation rate also capture complementary aspects of long-horizon behavior: models with similar average quality can exhibit substantially different temporal stability.

\noindent Fig.~\ref{fig:local-cases} grounds these aggregate trends in representative rollouts. Low-scoring examples exhibit visible degradation that closely follows the decline of the measured scores, often with failures emerging sharply near the end of a rollout. In contrast, several long rollouts sustain rich and coherent visual content throughout generation. These examples show both that long-horizon generation need not inevitably collapse and that the degradation captured by our local measures aligns with human-visible changes in generation quality.

\subsection{Long-Range Consistency}

\begin{wrapfigure}[17]{r}{0.5\textwidth}
\vspace{-8pt}
\setlength{\abovecaptionskip}{1pt}
\setlength{\belowcaptionskip}{0pt}
\captionsetup{width=\linewidth}
\centering
\includegraphics[width=\linewidth]{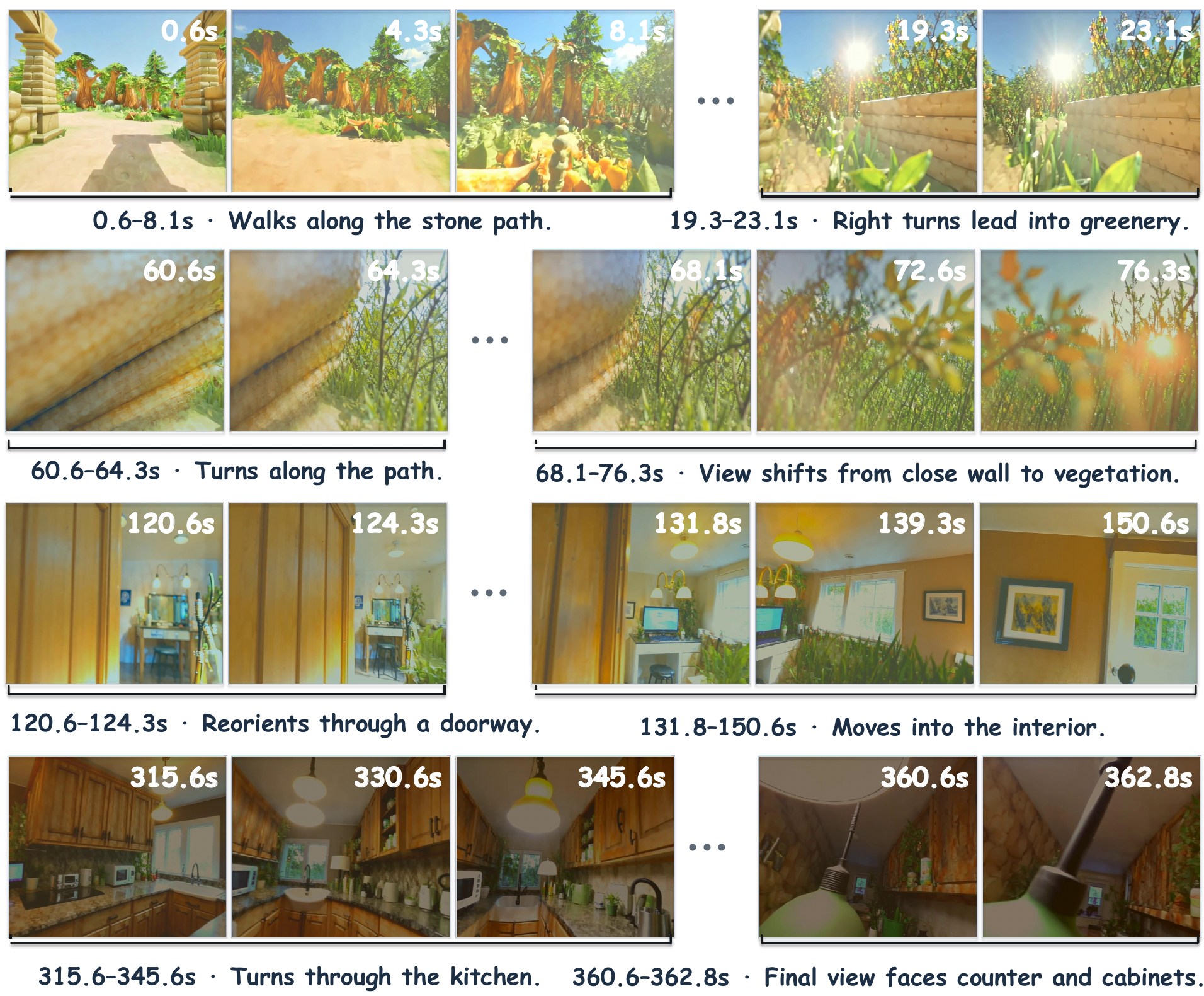}
\caption{\textbf{Extended closed-loop rollout.} Sampled observations from a 367.9\,s DreamX rollout driven by 491 agent actions.}
\label{fig:dreamx-agent-rollout}
\vspace{-4pt}
\end{wrapfigure}
Local degradation, however, captures only one side of long-horizon failure. We next examine inconsistencies that become visible only across extended spatial and temporal ranges. Table~\ref{tab:long-range} shows that long-range consistency remains weak across both spatial memory and temporal evolution, with most models scoring around or below the midpoint of the 0--10 scale.

\begingroup
\setlength{\floatsep}{4pt plus 1pt minus 2pt}
\setlength{\textfloatsep}{6pt plus 1pt minus 2pt}
\begin{figure}[!t]
\centering
\includegraphics[width=\linewidth]{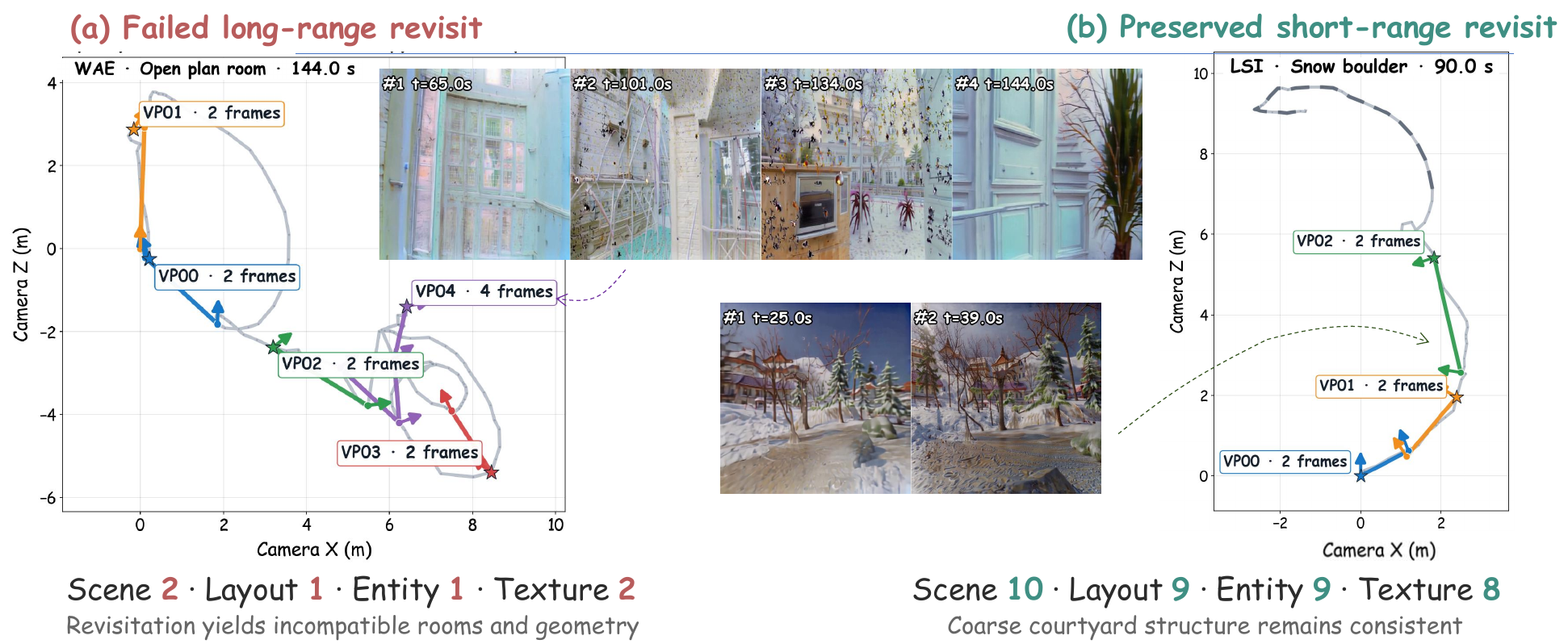}
\vspace{-6pt}
\setlength{\abovecaptionskip}{2pt}
\setlength{\belowcaptionskip}{0pt}
\caption{\textbf{Representative cases for spatial memory.} Long-range revisits show strong inconsistencies across spatially separated viewpoints, while short-range revisits better preserve scene structure; memory scores align with visible consistency across revisitation.}
\label{fig:memory-cases}
\end{figure}

\paragraph{Spatial memory.}
The memory results reveal a clear hierarchy in what current models can preserve. Under short occlusions or modest viewpoint changes, models often retain coarse scene layout while losing fine-grained texture, entity appearance, and local geometric details. The high-scoring example in Fig.~\ref{fig:memory-cases} illustrates this partial memory: the overall scene structure remains recognizable across revisits, while finer visual details drift.

This ability breaks down under larger spatial displacement or more substantial viewpoint changes. When the camera leaves a region for an extended period or revisits it from a different direction, models frequently generate content that is incompatible with the earlier observation, sometimes replacing the region with an entirely different scene. The low-scoring example in Fig.~\ref{fig:memory-cases} shows this behavior across multiple spatially distributed viewpoints. These results suggest that current models behave more like local re-renderers than systems with persistent global spatial memory: coarse structure may survive locally, but previously observed regions are not reliably preserved once interaction moves beyond the immediate neighborhood.

\begin{figure}[!t]
\centering
\includegraphics[width=\linewidth]{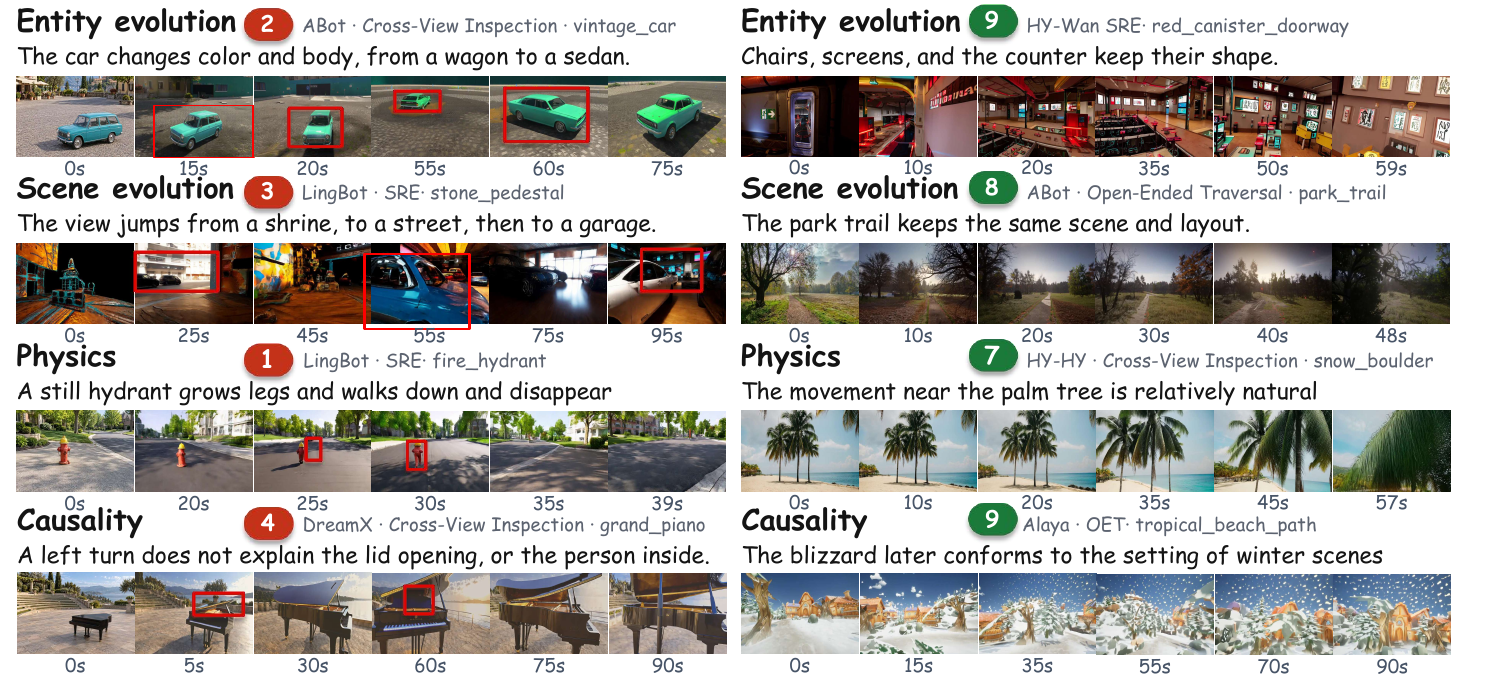}
\vspace{-6pt}
\setlength{\abovecaptionskip}{2pt}
\setlength{\belowcaptionskip}{0pt}
\caption{\textbf{Representative evolution behaviors.} Low-scoring cases exhibit entity and scene drift, implausible dynamics, and action-inconsistent changes, while high-scoring cases preserve substantially more coherent evolution.}
\label{fig:evolution-cases}
\end{figure}
\endgroup

\paragraph{Evolution fidelity.}
Spatial inconsistency is not limited to revisitation. Long rollouts also expose severe drift in how entities and scenes evolve over time. As illustrated in Fig.~\ref{fig:evolution-cases}, objects may deform, fragment, disappear, duplicate, or change identity, while scenes may gradually shift into inconsistent or unrelated content. These failures indicate that maintaining plausible individual frames is insufficient; a persistent world must also preserve the identity and state of its contents as interaction proceeds.

A second failure arises from incorrect attribution of visual change. Current models often conflate changes induced by camera motion, motion of visible objects, and intrinsic scene evolution. As a result, static objects may move or deform with the camera, scene changes may occur without corresponding actions, and dynamic behavior may violate basic physical or causal expectations. The evolution results therefore reveal a broader limitation: even when individual frames remain visually plausible, the model may fail to maintain a coherent account of what changed, why it changed, and which parts of the world should remain stable.

\begin{table*}[t]
\centering
\caption{\textbf{Long-range memory and evolution fidelity across models.} Memory evaluates scene, layout, entity, and texture consistency across revisited viewpoints; evolution evaluates entity, scene, physical, and causal consistency over extended rollouts.}
\label{tab:long-range}
\scriptsize
\setlength{\tabcolsep}{2.5pt}
\renewcommand{\arraystretch}{1.02}
\resizebox{\textwidth}{!}{%
\begin{tabular}{@{}l *{6}{r} *{6}{r} @{}}
\toprule
& \multicolumn{6}{c}{\textbf{Memory}} & \multicolumn{6}{c}{\textbf{Evolution}} \\
\cmidrule(lr){2-7}\cmidrule(lr){8-13}
\multicolumn{1}{c}{Model} & Scene & Layout & Entity & Texture & Overall & VP/seq. & Entity & Scene & Physics & Causality & Overall & Cxt/seq. \\
\midrule
Matrix-Game 2.0 & 4.73 & 3.66 & 3.47 & 4.19 & 4.01 & 2.14 & 3.78 & 3.59 & 3.60 & 3.28 & 3.56 & 12.68 \\
WorldPlay (HY) & 6.14 & 4.99 & 4.96 & 5.37 & 5.37 & 1.74 & 3.56 & 3.86 & 3.37 & 3.49 & 3.57 & 13.62 \\
WorldPlay (Wan) & 4.90 & 3.61 & 3.56 & 4.57 & 4.16 & 1.88 & 2.72 & 2.40 & 2.48 & 2.43 & 2.51 & 12.03 \\
LingBot-World & 4.59 & 3.22 & 2.92 & 4.37 & 3.78 & 2.66 & 4.30 & 3.79 & 3.77 & 3.21 & 3.77 & 12.23 \\
Matrix-Game 3.0 & 7.14 & 5.75 & 5.81 & 6.54 & 6.31 & 2.24 & 3.41 & 3.62 & 3.07 & 3.08 & 3.29 & 12.67 \\
SANA-WM & 6.54 & 4.96 & 5.07 & 5.91 & 5.62 & 2.13 & 4.30 & 4.40 & 3.81 & 3.74 & 4.06 & 10.25 \\
DreamX-World & 6.92 & 5.07 & 5.07 & 6.47 & 5.87 & 2.11 & \textbf{5.53} & \textbf{5.74} & \textbf{5.09} & \textbf{4.33} & \textbf{5.17} & 11.88 \\
AlayaWorld & \textbf{7.28} & \textbf{6.12} & \textbf{6.01} & \textbf{6.71} & \textbf{6.53} & 2.03 & 4.51 & 4.90 & 4.03 & 4.20 & 4.41 & 11.62 \\
LingBot-World v2 & 6.74 & 5.29 & 5.19 & 6.45 & 5.92 & 2.40 & 4.77 & 4.45 & 4.26 & 3.51 & 4.24 & 12.64 \\
ABot-World & 5.69 & 4.93 & 4.95 & 5.23 & 5.20 & 2.78 & 4.16 & 3.90 & 3.85 & 3.69 & 3.90 & 12.94 \\
\bottomrule
\end{tabular}%
}
\end{table*}

\subsection{Agent Effectiveness Analysis}
\label{sec:agent-effectiveness}

We further test whether the probing agent itself limits the achievable interaction horizon. In an unconstrained DreamX rollout (Fig.~\ref{fig:dreamx-agent-rollout}), the agent continues closed-loop exploration far beyond the 180\,s maximum horizon used in our benchmark and stops only after extended interaction. This suggests that the probing framework can support substantially longer evaluation horizons when the underlying video world model remains usable.

\section{Conclusion}

We introduce \textbf{WorldProbe}, a closed-loop benchmark for evaluating video world models over extended spatial and temporal horizons. Across 10 models, we find that local visual and content quality has become increasingly stable in recent systems, while spatial memory and evolution fidelity remain consistently weak. Overall, current video world models still fall short of maintaining a persistent world under long-horizon interaction.

\subsection*{AI Use Statement}

Generative AI tools were used to assist with code implementation and debugging, figure making and template preparation, and editing of the manuscript for clarity and presentation. All AI-assisted research decisions, code, experimental results, claims, figures, and manuscript content were reviewed and verified by the authors. The authors take full responsibility for the final content of this work.

\subsection*{Ethics Statement}

This work evaluates publicly available video world models and does not involve human subjects or personally identifiable data. We identify no additional ethical concerns specific to the proposed benchmark.

\subsection*{Reproducibility Statement}

We provide the task construction protocol, evaluation metrics, rollout settings, and model-specific implementation details in the main paper and appendix. Code, benchmark cases, and evaluation scripts are available at \url{https://github.com/Untitled-W/WorldProbe} to facilitate reproduction of our results.

\bibliography{references}

@misc{openai2024sora,
  title = {Video Generation Models as World Simulators},
  author = {{OpenAI}},
  year = {2024},
  url = {https://openai.com/research/video-generation-models-as-world-simulators}
}

@misc{wanteam2025wan,
  title = {Wan: Open and Advanced Large-Scale Video Generative Models},
  author = {{Team Wan} and Wang, Ang and others},
  year = {2025},
  eprint = {2503.20314},
  archivePrefix = {arXiv},
  primaryClass = {cs.CV},
  url = {https://arxiv.org/abs/2503.20314}
}

@misc{kong2024hunyuanvideo,
  title = {HunyuanVideo: A Systematic Framework For Large Video Generative Models},
  author = {Kong, Weijie and Tian, Qi and Zhang, Zijian and others},
  year = {2024},
  eprint = {2412.03603},
  archivePrefix = {arXiv},
  primaryClass = {cs.CV},
  url = {https://arxiv.org/abs/2412.03603}
}

@inproceedings{yang2024cogvideox,
  title = {{CogVideoX}: Text-to-Video Diffusion Models with An Expert Transformer},
  author = {Yang, Zhuoyi and Teng, Jiayan and Zheng, Wendi and others},
  booktitle = {International Conference on Learning Representations (ICLR)},
  year = {2025},
  eprint = {2408.06072},
  archivePrefix = {arXiv},
  primaryClass = {cs.CV},
  url = {https://proceedings.iclr.cc/paper_files/paper/2025/hash/ce31378e9f41d8907e97dab172b6c559-Abstract-Conference.html}
}

@inproceedings{bruce2024genie,
  title = {Genie: Generative Interactive Environments},
  author = {Bruce, Jake and Dennis, Michael and Edwards, Ashley and others},
  booktitle = {Proceedings of the 41st International Conference on Machine Learning},
  series = {Proceedings of Machine Learning Research},
  volume = {235},
  pages = {4603--4623},
  publisher = {PMLR},
  year = {2024},
  eprint = {2402.15391},
  archivePrefix = {arXiv},
  primaryClass = {cs.LG},
  url = {https://proceedings.mlr.press/v235/bruce24a.html}
}

@misc{deepmind2024genie2,
  title = {{Genie 2}: A large-scale foundation world model},
  author = {Parker-Holder, Jack and Ball, Philip and Bruce, Jake and Dasagi, Vibhavari and Holsheimer, Kristian and Kaplanis, Christos and Moufarek, Alexandre and Scully, Guy and Shar, Jeremy and Shi, Jimmy and Spencer, Stephen and Yung, Jessica and Dennis, Michael and Kenjeyev, Sultan and Long, Shangbang and Mnih, Vlad and Chan, Harris and Gazeau, Maxime and Li, Bonnie and Pardo, Fabio and Wang, Luyu and Zhang, Lei and Besse, Frederic and Harley, Tim and Mitenkova, Anna and Wang, Jane and Clune, Jeff and Hassabis, Demis and Hadsell, Raia and Bolton, Adrian and Singh, Satinder and Rockt{\"a}schel, Tim},
  year = {2024},
  url = {https://deepmind.google/blog/genie-2-a-large-scale-foundation-world-model/}
}

@misc{deepmind2025genie3,
  title = {Genie 3: A new frontier for world models},
  author = {Parker-Holder, Jack and Fruchter, Shlomi},
  year = {2025},
  url = {https://deepmind.google/blog/genie-3-a-new-frontier-for-world-models/}
}

@misc{zhang2025matrixgame,
  title = {Matrix-Game: Interactive World Foundation Model},
  author = {Zhang, Yifan and Peng, Chunli and Wang, Boyang and others},
  year = {2025},
  eprint = {2506.18701},
  archivePrefix = {arXiv},
  primaryClass = {cs.CV},
  url = {https://arxiv.org/abs/2506.18701}
}

@misc{li2025hunyuan_gamecraft,
  title = {Hunyuan-GameCraft: High-dynamic Interactive Game Video Generation with Hybrid History Condition},
  author = {Li, Jiaqi and Tang, Junshu and Xu, Zhiyong and others},
  year = {2025},
  eprint = {2506.17201},
  archivePrefix = {arXiv},
  primaryClass = {cs.CV},
  url = {https://arxiv.org/abs/2506.17201}
}

@misc{wang2026matrixgame3,
  title = {Matrix-Game 3.0: Real-Time and Streaming Interactive World Model with Long-Horizon Memory},
  author = {Wang, Zile and Liu, Zexiang and Li, Jiaxing and others},
  year = {2026},
  eprint = {2604.08995},
  archivePrefix = {arXiv},
  primaryClass = {cs.CV},
  url = {https://arxiv.org/abs/2604.08995}
}

@misc{tang2025hunyuan_gamecraft2,
  title = {Hunyuan-GameCraft-2: Instruction-following Interactive Game World Model},
  author = {Tang, Junshu and Liu, Jiacheng and Li, Jiaqi and others},
  year = {2025},
  eprint = {2511.23429},
  archivePrefix = {arXiv},
  primaryClass = {cs.CV},
  url = {https://arxiv.org/abs/2511.23429}
}

@misc{mao2025yume,
  title = {Yume: An Interactive World Generation Model},
  author = {Mao, Xiaofeng and others},
  year = {2025},
  eprint = {2507.17744},
  archivePrefix = {arXiv},
  primaryClass = {cs.CV},
  url = {https://arxiv.org/abs/2507.17744}
}

@misc{mao2025yume15,
  title = {Yume-1.5: A Text-Controlled Interactive World Generation Model},
  author = {Mao, Xiaofeng and Li, Zhen and Li, Chuanhao and others},
  year = {2025},
  eprint = {2512.22096},
  archivePrefix = {arXiv},
  primaryClass = {cs.CV},
  url = {https://arxiv.org/abs/2512.22096}
}

@misc{gao2026lingbotworld,
  title = {Advancing Open-source World Models},
  author = {{Robbyant Team} and Gao, Zelin and Wang, Qiuyu and Zeng, Yanhong and others},
  year = {2026},
  eprint = {2601.20540},
  archivePrefix = {arXiv},
  primaryClass = {cs.CV},
  url = {https://arxiv.org/abs/2601.20540}
}

@misc{wu2026infiniteworld,
  title = {Infinite-World: Scaling Interactive World Models to 1000-Frame Horizons via Pose-Free Hierarchical Memory},
  author = {Wu, Ruiqi and He, Xuanhua and Cheng, Meng and others},
  year = {2026},
  eprint = {2602.02393},
  archivePrefix = {arXiv},
  primaryClass = {cs.CV},
  url = {https://arxiv.org/abs/2602.02393}
}

@inproceedings{sun2025worldplay,
  title = {{WorldPlay}: Towards Long-Term Geometric Consistency for Real-Time Interactive World Modeling},
  author = {Sun, Wenqiang and Zhang, Haiyu and Wang, Haoyuan and others},
  booktitle = {Forty-third International Conference on Machine Learning},
  year = {2026},
  eprint = {2512.14614},
  archivePrefix = {arXiv},
  primaryClass = {cs.CV},
  url = {https://openreview.net/forum?id=GfSwkDSr8J}
}

@inproceedings{ren2025gen3c,
  title = {{GEN3C}: {3D}-Informed World-Consistent Video Generation with Precise Camera Control},
  author = {Ren, Xuanchi and others},
  booktitle = {Proceedings of the IEEE/CVF Conference on Computer Vision and Pattern Recognition (CVPR)},
  pages = {6121--6132},
  year = {2025},
  doi = {10.1109/CVPR52734.2025.00574},
  eprint = {2503.03751},
  archivePrefix = {arXiv},
  primaryClass = {cs.CV},
  url = {https://openaccess.thecvf.com/content/CVPR2025/html/Ren_GEN3C_3D-Informed_World-Consistent_Video_Generation_with_Precise_Camera_Control_CVPR_2025_paper.html}
}

@misc{hunyuanworld2025,
  title = {HunyuanWorld 1.0: Generating Immersive, Explorable, and Interactive 3D Worlds from Words or Pixels},
  author = {{HunyuanWorld Team} and Wang, Zhenwei and Liu, Yuhao and Wu, Junta and others},
  year = {2025},
  eprint = {2507.21809},
  archivePrefix = {arXiv},
  primaryClass = {cs.CV},
  url = {https://arxiv.org/abs/2507.21809}
}

@misc{yang2025matrix3d,
  title = {{Matrix-3D}: Omnidirectional Explorable {3D} World Generation},
  author = {Yang, Zhongqi and Ge, Wenhang and Li, Yuqi and others},
  year = {2025},
  eprint = {2508.08086},
  archivePrefix = {arXiv},
  primaryClass = {cs.CV},
  url = {https://arxiv.org/abs/2508.08086}
}

@misc{alayaworld2026,
  title = {AlayaWorld: Interactive Long-Horizon World Modeling -- Full Technical Report},
  author = {{AlayaWorld Team} and Zhang, Kaipeng and Li, Chuanhao and others},
  year = {2026},
  eprint = {2607.18367},
  archivePrefix = {arXiv},
  primaryClass = {cs.CV},
  url = {https://arxiv.org/abs/2607.18367}
}

@misc{lingbotworld22026,
  title = {Infinite Worlds with Versatile Interactions},
  author = {Gao, Zelin and Wang, Qiuyu and Zhu, Jiapeng and others},
  year = {2026},
  eprint = {2607.07534},
  archivePrefix = {arXiv},
  primaryClass = {cs.CV},
  url = {https://arxiv.org/abs/2607.07534}
}

@misc{jiang2026abotworld,
  title = {{ABot-World-0}: Infinite Interactive World Rollout on a Single Desktop {GPU}},
  author = {Jiang, Fan and Sun, Zhaoxu and Wang, Mengchao and others},
  year = {2026},
  eprint = {2607.19191},
  archivePrefix = {arXiv},
  primaryClass = {cs.CV},
  url = {https://arxiv.org/abs/2607.19191}
}

@misc{dreamxworld2026,
  title = {DreamX-World 1.0: A General-Purpose Interactive World Model},
  author = {{DreamX Team} and Bai, Yancheng and Chen, Rui and Chu, Xiangxiang and Dang, Rujing and Dou, Hao and Gao, Bingjie and Gu, Qiwen and Hong, Siyu and Lei, Jiachen and Li, Geng and Li, Jifan and Lin, Ruimin and Shi, Qingfeng and Song, Bingze and Sun, Lei and Tang, Jing and Tian, Ruitian and Wang, Jun and Wu, Jiahong and Zhang, Pengfei and Zhang, Shen and Zhu, Jiashu},
  year = {2026},
  eprint = {2606.16993},
  archivePrefix = {arXiv},
  primaryClass = {cs.CV},
  url = {https://arxiv.org/abs/2606.16993}
}

@misc{he2025matrixgame2,
  title = {{Matrix-Game} 2.0: An Open-Source, Real-Time, and Streaming Interactive World Model},
  author = {He, Xianglong and Peng, Chunli and Liu, Zexiang and others},
  year = {2025},
  eprint = {2508.13009},
  archivePrefix = {arXiv},
  primaryClass = {cs.CV},
  url = {https://arxiv.org/abs/2508.13009}
}

@misc{zhu2026sanawm,
  title = {{SANA-WM}: Efficient Minute-Scale World Modeling with Hybrid Linear Diffusion Transformer},
  author = {Zhu, Haoyi and Liu, Haozhe and Zhao, Yuyang and others},
  year = {2026},
  eprint = {2605.15178},
  archivePrefix = {arXiv},
  primaryClass = {cs.CV},
  url = {https://arxiv.org/abs/2605.15178}
}

@misc{qwen2026qwen36,
  title = {{Qwen3.6-27B}},
  author = {{Qwen Team}},
  year = {2026},
  url = {https://huggingface.co/Qwen/Qwen3.6-27B}
}

@misc{carion2025sam3,
  title = {{SAM} 3: Segment Anything with Concepts},
  author = {Carion, Nicolas and Gustafson, Laura and Hu, Yuan-Ting and others},
  year = {2025},
  eprint = {2511.16719},
  archivePrefix = {arXiv},
  primaryClass = {cs.CV},
  url = {https://arxiv.org/abs/2511.16719}
}

@misc{lin2025depthanything3,
  title = {Depth Anything 3: Recovering the Visual Space from Any Views},
  author = {Lin, Haotong and Chen, Sili and Liew, Junhao and others},
  year = {2025},
  eprint = {2511.10647},
  archivePrefix = {arXiv},
  primaryClass = {cs.CV},
  url = {https://arxiv.org/abs/2511.10647}
}

@misc{matrixgame352026,
  title = {{Matrix-Game} 3.5: Enhancing Real-Time Streaming Interactive World Models with Patch Memory},
  author = {Qian, Runjia and Wang, Zile and Zhang, Jihai and others},
  year = {2026},
  eprint = {2608.29910},
  archivePrefix = {arXiv},
  primaryClass = {cs.CV},
  url = {https://arxiv.org/abs/2608.29910}
}

@misc{wonder2026,
  title = {Wonder: Video World Model Done Better},
  author = {Xu, Jiacong and Jiang, Hanwen and Shu, Zhixin and Sunkavalli, Kalyan and Patel, Vishal M. and Mei, Yiqun},
  year = {2026},
  eprint = {2607.26037},
  archivePrefix = {arXiv},
  primaryClass = {cs.CV},
  url = {https://arxiv.org/abs/2607.26037}
}

@misc{gimworld2026,
  title = {Geometry-Aware Implicit Memory for Video World Models},
  author = {Wei, Zhengxuan and Guo, Xu and Li, Xinghui and others},
  year = {2026},
  eprint = {2606.02436},
  archivePrefix = {arXiv},
  primaryClass = {cs.CV},
  url = {https://arxiv.org/abs/2606.02436}
}

@misc{mirage2026,
  title = {Latent Spatial Memory for Video World Models},
  author = {Wang, Weijie and Zhao, Haoyu and Yang, Yifan and others},
  year = {2026},
  eprint = {2606.09828},
  archivePrefix = {arXiv},
  primaryClass = {cs.CV},
  url = {https://arxiv.org/abs/2606.09828}
}

@misc{worlddirector2026,
  title = {{WorldDirector}: Building Controllable World Simulators with Persistent Dynamic Memory},
  author = {Wang, Hanlin and Ouyang, Hao and Wang, Qiuyu and others},
  year = {2026},
  eprint = {2607.02517},
  archivePrefix = {arXiv},
  primaryClass = {cs.CV},
  url = {https://arxiv.org/abs/2607.02517}
}

@misc{worldinworld2026,
  title = {World in World: Explore the World with World Models},
  author = {Song, Chenxi and Yang, Yanming and Zhang, Chi},
  year = {2026},
  eprint = {2609.11548},
  archivePrefix = {arXiv},
  primaryClass = {cs.CV},
  url = {https://arxiv.org/abs/2609.11548}
}

@inproceedings{huang2023vbench,
  title = {{VBench}: Comprehensive Benchmark Suite for Video Generative Models},
  author = {Huang, Ziqi and He, Yinan and Yu, Jiashuo and others},
  booktitle = {Proceedings of the IEEE/CVF Conference on Computer Vision and Pattern Recognition (CVPR)},
  pages = {21807--21818},
  year = {2024},
  doi = {10.1109/CVPR52733.2024.02060},
  eprint = {2311.17982},
  archivePrefix = {arXiv},
  primaryClass = {cs.CV},
  url = {https://openaccess.thecvf.com/content/CVPR2024/html/Huang_VBench_Comprehensive_Benchmark_Suite_for_Video_Generative_Models_CVPR_2024_paper.html}
}

@article{huang2024vbenchpp,
  title = {{VBench++}: Comprehensive and Versatile Benchmark Suite for Video Generative Models},
  author = {Huang, Ziqi and Zhang, Fan and Xu, Xiaojie and others},
  journal = {IEEE Transactions on Pattern Analysis and Machine Intelligence},
  volume = {48},
  number = {3},
  pages = {3268--3285},
  year = {2026},
  month = mar,
  doi = {10.1109/TPAMI.2025.3633890},
  eprint = {2411.13503},
  archivePrefix = {arXiv},
  primaryClass = {cs.CV},
  url = {https://doi.org/10.1109/TPAMI.2025.3633890}
}

@misc{zheng2025vbench2,
  title = {{VBench-2.0}: Advancing Video Generation Benchmark Suite for Intrinsic Faithfulness},
  author = {Zheng, Dian and Huang, Ziqi and Liu, Hongbo and others},
  year = {2025},
  eprint = {2503.21755},
  archivePrefix = {arXiv},
  primaryClass = {cs.CV},
  url = {https://arxiv.org/abs/2503.21755}
}

@inproceedings{sun2024t2vcompbench,
  title = {{T2V-CompBench}: A Comprehensive Benchmark for Compositional Text-to-video Generation},
  author = {Sun, Kaiyue and others},
  booktitle = {Proceedings of the IEEE/CVF Conference on Computer Vision and Pattern Recognition (CVPR)},
  pages = {8406--8416},
  year = {2025},
  eprint = {2407.14505},
  archivePrefix = {arXiv},
  primaryClass = {cs.CV},
  url = {https://openaccess.thecvf.com/content/CVPR2025/html/Sun_T2V-CompBench_A_Comprehensive_Benchmark_for_Compositional_Text-to-video_Generation_CVPR_2025_paper.html}
}

@inproceedings{li2025worldmodelbench,
  title = {{WorldModelBench}: Judging Video Generation Models As World Models},
  author = {Li, Dacheng and Fang, Yunhao and Chen, Yukang and others},
  booktitle = {Advances in Neural Information Processing Systems},
  volume = {38},
  pages = {61064--61088},
  year = {2025},
  note = {Datasets and Benchmarks Track},
  doi = {10.52202/085713-1834},
  eprint = {2502.20694},
  archivePrefix = {arXiv},
  primaryClass = {cs.CV},
  url = {https://proceedings.neurips.cc/paper_files/paper/2025/hash/4ec03ed08a3fcb59e1c815b5598beff1-Abstract-Datasets_and_Benchmarks_Track.html}
}

@inproceedings{qin2024worldsimbench,
  title = {{WorldSimBench}: Towards Video Generation Models as World Simulators},
  author = {Qin, Yiran and Shi, Zhelun and Yu, Jiwen and others},
  booktitle = {Proceedings of the 42nd International Conference on Machine Learning},
  series = {Proceedings of Machine Learning Research},
  volume = {267},
  pages = {50338--50362},
  publisher = {PMLR},
  year = {2025},
  eprint = {2410.18072},
  archivePrefix = {arXiv},
  primaryClass = {cs.CV},
  url = {https://proceedings.mlr.press/v267/qin25f.html}
}

@inproceedings{duan2025worldscore,
  title = {{WorldScore}: A Unified Evaluation Benchmark for World Generation},
  author = {Duan, Haoyi and Yu, Hong-Xing and Chen, Sirui and Li, Fei-Fei and Wu, Jiajun},
  booktitle = {Proceedings of the IEEE/CVF International Conference on Computer Vision (ICCV)},
  pages = {27713--27724},
  year = {2025},
  doi = {10.1109/ICCV51701.2025.02573},
  eprint = {2504.00983},
  archivePrefix = {arXiv},
  primaryClass = {cs.CV},
  url = {https://openaccess.thecvf.com/content/ICCV2025/html/Duan_WorldScore_A_Unified_Evaluation_Benchmark_for_World_Generation_ICCV_2025_paper.html}
}

@misc{wu2026omniworldbench,
  title = {Omni-WorldBench: Towards a Comprehensive Interaction-Centric Evaluation for World Models},
  author = {Wu, Meiqi and Cai, Zhixin and Zhao, Fufangchen and others},
  year = {2026},
  eprint = {2603.22212},
  archivePrefix = {arXiv},
  primaryClass = {cs.CV},
  url = {https://arxiv.org/abs/2603.22212}
}

@misc{worldmark2026,
  title = {WorldMark: A Unified Benchmark Suite for Interactive Video World Models},
  author = {Xu, Xiaojie and Lin, Zhengyuan and He, Kang and Feng, Yukang and Mao, Xiaofeng and Yin, Yuanyang and Ge, Yongtao and Zhang, Kaipeng},
  year = {2026},
  eprint = {2604.21686},
  archivePrefix = {arXiv},
  primaryClass = {cs.CV},
  url = {https://arxiv.org/abs/2604.21686}
}

@misc{fang2026iworldbench,
  title = {iWorld-Bench: A Benchmark for Interactive World Models with a Unified Action Generation Framework},
  author = {Fang, Jianjie and Lei, Yingshan and Wan, Qin and others},
  year = {2026},
  eprint = {2605.03941},
  archivePrefix = {arXiv},
  primaryClass = {cs.CV},
  url = {https://arxiv.org/abs/2605.03941}
}

@misc{wbench2026,
  title = {WBench: A Comprehensive Multi-turn Benchmark for Interactive Video World Model Evaluation},
  author = {Ying, Kaining and Hu, Hengrui and Ren, Siyu and Li, Jiamu and Chen, Fengjiao and Wang, Ziwen and Cao, Xuezhi and Cai, Xunliang and Ding, Henghui},
  year = {2026},
  eprint = {2605.25874},
  archivePrefix = {arXiv},
  primaryClass = {cs.CV},
  url = {https://arxiv.org/abs/2605.25874}
}

@misc{lian2025loopnav,
  title = {LoopNav: Benchmarking Spatial Consistency in World Models},
  author = {Lian, Kewei and Cai, Shaofei and Liang, Yitao and Liu, Anji},
  year = {2025},
  eprint = {2505.22976},
  archivePrefix = {arXiv},
  primaryClass = {cs.CV},
  url = {https://arxiv.org/abs/2505.22976}
}

@misc{ye2026mind,
  title = {MIND: Benchmarking Memory Consistency and Action Control in World Models},
  author = {Ye, Yixuan and Lu, Xuanyu and Jiang, Yuxin and Gu, Yuchao and Zhao, Rui and Liang, Qiwei and Pan, Jiachun and Zhang, Fengda and Wu, Weijia and Wang, Alex Jinpeng},
  year = {2026},
  eprint = {2602.08025},
  archivePrefix = {arXiv},
  primaryClass = {cs.CV},
  url = {https://arxiv.org/abs/2602.08025}
}

@misc{zhang2026mbench,
  title = {MBench: A Comprehensive Benchmark on Memory Capability for Video World Models},
  author = {Zhang, Shengjun and Zhang, Zhang and Huang, Simin and Tang, Zhenyu and Wang, Hanyang and Dai, Chensheng and Chen, Min and Li, Yifan and Li, Yuxin and Chen, Yingjie and Liu, Hao and Li, Chen and Lyu, Jing and Duan, Yueqi},
  year = {2026},
  eprint = {2606.00793},
  archivePrefix = {arXiv},
  primaryClass = {cs.CV},
  url = {https://arxiv.org/abs/2606.00793}
}

@misc{huang2025vipe,
  title = {{ViPE}: Video Pose Engine for {3D} Geometric Perception},
  author = {Huang, Jiahui and Zhou, Qunjie and Rabeti, Hesam and Korovko, Aleksandr and Ling, Huan and Ren, Xuanchi and Shen, Tianchang and Gao, Jun and Slepichev, Dmitry and Lin, Chen-Hsuan and Ren, Jiawei and Xie, Kevin and Biswas, Joydeep and Leal-Taixe, Laura and Fidler, Sanja},
  year = {2025},
  eprint = {2508.10934},
  archivePrefix = {arXiv},
  primaryClass = {cs.CV},
  url = {https://arxiv.org/abs/2508.10934}
}
\bibliographystyle{iclr2027_conference}

\appendix
\section{Local Metrics by Dimension}
\label{app:local-metrics-dimensions}

Table~\ref{tab:local-metrics-dimensions} reports the per-dimension visual-IQA and content scores underlying the overall aggregates in Table~\ref{tab:local-metrics}.

\begin{table*}[t]
\centering
\caption{Per-dimension frame-level visual-IQA and content scores across the 327 rollouts with complete scores in both groups. For each model and dimension, Mean is the equal-weighted average of rollout-level frame means; Slope is the equal-weighted average of per-rollout least-squares slopes against elapsed time in seconds. Visual scores are normalized to their theoretical ranges (MUSIQ: 0--100, MANIQA: 0--1, Q-Align: 1--5); their slopes are reported in $10^{-3}$/s. Content dimensions retain their 0--10 scale and slopes are in points/s. Best mean and slope in each column are in bold.}
\label{tab:local-metrics-dimensions}
\scriptsize
\setlength{\tabcolsep}{2pt}
\renewcommand{\arraystretch}{1.02}
\resizebox{\textwidth}{!}{%
\begin{tabular}{@{}l *{12}{r}@{}}
\toprule
\multirow{3}{*}{Model} & \multicolumn{6}{c}{Visual (normalized)} & \multicolumn{6}{c}{Content (0--10)} \\
\cmidrule(lr){2-7}\cmidrule(lr){8-13}
& \multicolumn{2}{c}{MUSIQ} & \multicolumn{2}{c}{MANIQA} & \multicolumn{2}{c}{Q-Align} & \multicolumn{2}{c}{\shortstack{Geometry}} & \multicolumn{2}{c}{\shortstack{Object\\Validity}} & \multicolumn{2}{c}{\shortstack{Scene\\Richness}} \\
\cmidrule(lr){2-3}\cmidrule(lr){4-5}\cmidrule(lr){6-7}\cmidrule(lr){8-9}\cmidrule(lr){10-11}\cmidrule(lr){12-13}
& Mean & Slope & Mean & Slope & Mean & Slope & Mean & Slope & Mean & Slope & Mean & Slope \\
\midrule
\mbox{Matrix-Game 2.0} & 0.667 & $-0.08$ & 0.378 & $-0.22$ & 0.649 & $-2.01$ & 6.14 & $-0.060$ & 6.07 & $-0.063$ & 6.07 & $-0.059$ \\
\mbox{WorldPlay (HY)} & 0.622 & $-1.87$ & 0.346 & $-1.13$ & 0.657 & $-6.89$ & 5.97 & $-0.077$ & 5.90 & $-0.082$ & 6.38 & $-0.057$ \\
\mbox{WorldPlay (Wan)} & \textbf{0.687} & $-0.24$ & \textbf{0.400} & $-0.92$ & 0.722 & $-7.57$ & 6.06 & $-0.102$ & 5.94 & $-0.117$ & 6.87 & $-0.058$ \\
\mbox{LingBot-World} & 0.606 & $-1.70$ & 0.331 & $-2.21$ & 0.660 & $-2.90$ & 6.88 & $-0.021$ & 6.66 & $-0.031$ & 6.28 & $-0.026$ \\
\mbox{Matrix-Game 3.0} & 0.613 & \textbf{$+1.65$} & 0.346 & \textbf{$+1.69$} & 0.708 & $-1.06$ & 5.70 & $-0.055$ & 5.63 & $-0.059$ & 5.95 & $-0.039$ \\
\mbox{SANA-WM} & 0.508 & $-2.64$ & 0.231 & $-2.35$ & 0.682 & $-4.98$ & 6.68 & $-0.068$ & 6.78 & $-0.072$ & 7.28 & $-0.023$ \\
\mbox{DreamX-World} & 0.482 & $-3.55$ & 0.227 & $-2.37$ & 0.664 & $-3.40$ & 7.31 & $-0.023$ & 7.39 & $-0.022$ & \textbf{7.37} & \textbf{$-0.009$} \\
\mbox{AlayaWorld} & 0.618 & $-2.45$ & 0.362 & $+0.11$ & 0.699 & $-5.50$ & 6.34 & $-0.088$ & 6.15 & $-0.103$ & 5.92 & $-0.089$ \\
\mbox{LingBot-World v2} & 0.639 & $-1.01$ & 0.362 & $-1.26$ & \textbf{0.748} & \textbf{$-0.78$} & \textbf{7.75} & \textbf{$-0.015$} & \textbf{7.83} & \textbf{$-0.015$} & 7.08 & $-0.015$ \\
\mbox{ABot-World} & 0.464 & $-5.60$ & 0.261 & $-2.32$ & 0.591 & $-8.52$ & 6.47 & $-0.061$ & 6.28 & $-0.073$ & 6.43 & $-0.062$ \\
\bottomrule
\end{tabular}%
}
\end{table*}

\section{Implementation Details}
\label{app:implementation_details}

\subsection{Service Architecture and Rollout Lifecycle}
\label{app:system_architecture}

WorldProbe separates task control from model inference. The experiment runner owns each rollout's task state, observation history, action budget, and termination checks. At each decision point, it prepares the task instructions and selected observations for the task agent. The agent makes a structured decision with the Qwen3.6-27B VLM service and returns a discrete camera action. SAM3.1 maintains the target mask and identity across generated chunks, while Depth-Anything 3 (DA3) supplies local depth for near-field collision checks. The runner combines these signals with the agent decision and sends the resulting command to the resident world-model service.

The World Model Interface (WMI) exposes HTTP session operations and keeps model weights resident. A model-specific driver translates each command into the backend's native control representation and advances one native generation unit while retaining its latent, attention-cache, scheduler, and decoder state. WMI consumes the returned frame chunks locally, writes preview and final video artifacts, and returns compact metadata to the runner. The runner then prepares the next observation and repeats the loop. After generation, rollout screening and metric evaluation operate on the saved artifacts; raw GPU tensors are not sent through the HTTP API.

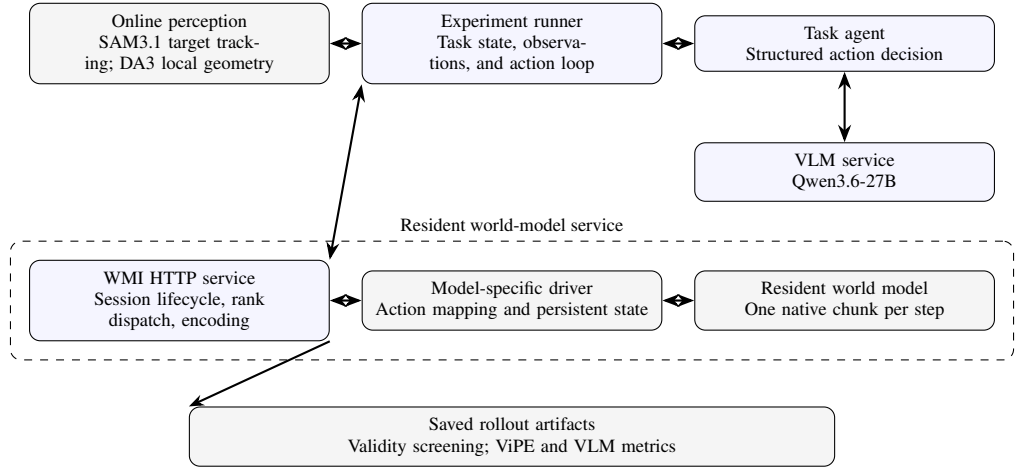
\begin{figure}[t]
\centering
\begin{tikzpicture}[
  component/.style={draw, rounded corners, align=center, inner sep=4pt,
                    minimum height=0.72cm, font=\scriptsize},
  service/.style={component, fill=blue!4},
  module/.style={component, fill=gray!8},
  link/.style={<->, >=Stealth, thick},
  flow/.style={-{Stealth[length=2mm]}, thick}
]
\node[module, text width=1.45in] (perception) at (0,1.8)
  {Online perception\\SAM3.1 target tracking; DA3 local geometry};
\node[service, text width=1.45in] (runner) at (4.4,1.8)
  {Experiment runner\\Task state, observations, and action loop};
\node[service, text width=1.45in] (agent) at (8.8,1.8)
  {Task agent\\Structured action decision};
\node[service, text width=1.45in] (vlm) at (8.8,0.1)
  {VLM service\\Qwen3.6-27B};

\node[service, text width=1.45in] (wmi) at (0,-1.6)
  {WMI HTTP service\\Session lifecycle, rank dispatch, encoding};
\node[module, text width=1.45in] (driver) at (4.4,-1.6)
  {Model-specific driver\\Action mapping and persistent state};
\node[module, text width=1.45in] (model) at (8.8,-1.6)
  {Resident world model\\One native chunk per step};
\node[module, text width=3.25in] (evaluation) at (4.4,-3.4)
  {Saved rollout artifacts\\Validity screening; ViPE and VLM metrics};

\node[draw, dashed, rounded corners, inner sep=7pt,
      fit=(wmi)(driver)(model),
      label={[font=\scriptsize]above:Resident world-model service}] {};

\draw[link] (perception.east) -- (runner.west);
\draw[link] (runner.east) -- (agent.west);
\draw[link] (agent.south) -- (vlm.north);
\draw[link] (runner.south west) -- (wmi.north east);
\draw[link] (wmi.east) -- (driver.west);
\draw[link] (driver.east) -- (model.west);
\draw[flow] (wmi.south east) -- (evaluation.north west);
\end{tikzpicture}
\caption{WorldProbe's service and data flow. The runner mediates the task agent, auxiliary perception modules, and WMI. Bidirectional arrows indicate request and response; the final artifact flows to rollout screening and metric evaluation.}
\label{fig:implementation-architecture}
\end{figure}

The rollout ends when the agent declares the current spatial objective complete, when tracking failure or oscillatory control persists, or when the task budget is exhausted. Before scoring, we apply minimum-duration checks and coarse VLM screening for severe visual or trajectory anomalies. Attempts below the task-specific duration threshold are retried up to three times, retaining the longest attempt. WMI's finalizer writes the accumulated native chunks as one video artifact, independently of the step previews.

\subsection{Model-specific Drivers and Streaming State}
\label{app:driver-details}

Each driver implements the same six lifecycle operations: load and close a resident model, start and finish a session, advance one step, and abort a session. The step boundary follows each model's native generation unit; the driver retains the backend-specific state between calls. Table~\ref{tab:driver-config} reports GPU allocation, code-derived session capacities, and state parameters. The configured chunk capacities are session bounds, not the lengths of individual evaluated rollouts. This table describes chunk and cache settings, rather than repeating the main paper's runtime columns.

\begin{table}[t]
\centering
\caption{Code-derived streaming capacities and model-specific state.}
\label{tab:driver-config}
\small
\setlength{\tabcolsep}{3pt}
\renewcommand{\arraystretch}{1.08}
\begin{tabularx}{\linewidth}{@{}l>{\centering\arraybackslash}p{0.45in}>{\centering\arraybackslash}p{0.72in}X@{}}
\toprule
\multicolumn{1}{c}{Model / driver} & \multicolumn{1}{c}{GPUs} & \multicolumn{1}{c}{\shortstack{Configured\\chunk bound}} & \multicolumn{1}{>{\centering\arraybackslash}X}{Native step, cache, and adaptation} \\
\midrule
Matrix-Game 2.0 & 1 & 180 & 3 latents per action; a 4- or 6-latent visual-KV window by game; 32-slot causal-VAE feature cache; the token cursor advances by 2,640 per step. \\
WorldPlay (Hunyuan) & 2 & 270 & 4 latents per action; geometry-aware history retrieval from up to 20 frames. \\
WorldPlay (Wan) & 4 & 180 & 4 latents per action; a 16-latent visual context composed of 12 recent and 4 field-of-view-selected remote latents. \\
LingBot-World (v1 Fast) & 2 & 240 & 3 latents per action; local attention 18 with 6 sink latents; four denoising evaluations per chunk. \\
Matrix-Game 3.0 & 2 & 77 & Initial segment of 15 latents; subsequent segments add 10 latents with 4 preceding latents for image conditioning; field-of-view memory selects 5 latents. \\
SANA-WM & 1 & 120 & 24 pose updates per action; 2 cached Stage-1 blocks; refiner block size 3 and KV limit 11 frames; denoising schedule $[1000,960,889,727,0]$. \\
DreamX-World & 1 & 240 & 3 latents per action; self- and cross-attention caches across 30 layers; four denoising steps per chunk. \\
AlayaWorld & 2 & 135 & 4 latents per action; camera history uses 33 poses at temporal stride 8; each chunk encodes 2.54 m translation or 22.5$^\circ$ rotation. \\
LingBot-World v2 & 2 & 180 & 4 latents per action; an 18-slot self-KV cache with 6 sink and 12 rolling slots. \\
ABot-World & 1 & 180 & 3 latents per action; four denoising updates; the initial frame uses a center-cover crop to $1280\times704$. \\
\bottomrule
\end{tabularx}
\end{table}

The driver leaves model-specific denoising, action conditioning, and cache updates in the upstream model components. It externalizes the chunk boundary so that the runner can choose the next action after seeing the preceding output. For example, the Matrix-Game 2 and ABot drivers preserve their causal VAE feature cache across steps; the LingBot drivers preserve self-KV, text cross-KV, and decoder state; and the Hunyuan driver retains text KV while rebuilding visual KV from the history selected for the current camera pose. WMI owns the common HTTP protocol, rank dispatch, video encoding, artifact paths, and telemetry, rather than duplicating those responsibilities in each driver.

\clearpage
\subsection{Public Inference Paths and GPU-Memory Constraints}
\label{app:upstream-constraints}

We inspected the public inference entrypoints and reproduced their native chunk or segment loops where available. Table~\ref{tab:public-paths} gives the largest directly recorded public-entrypoint run for each evaluated backend and converts its decoded frame count to video duration. These are observed run lengths, not claimed hard limits of the model architecture. They expose the distinction between generating chunks inside a fixed invocation and making each chunk available to an external controller before the next action is chosen.

\begin{table}[H]
\centering
\caption{Recorded horizons and control or memory behavior of the public inference paths. Durations are computed from output frames and the corresponding public-path frame rate.}
\label{tab:public-paths}
\footnotesize
\setlength{\tabcolsep}{3pt}
\renewcommand{\arraystretch}{1.02}
\begin{tabularx}{\linewidth}{@{}l>{\centering\arraybackslash}p{1.18in}X@{}}
\toprule
\multicolumn{1}{c}{Model} & \multicolumn{1}{c}{\shortstack{Largest recorded\\native run}} & \multicolumn{1}{>{\centering\arraybackslash}X}{Public-path control and memory behavior} \\
\midrule
Matrix-Game 2.0 & 10 blocks; 117 frames, 9.75 s & 3-latent loop with rolling visual/action KV and persistent causal-VAE cache; each block waits for an action on stdin. The standalone entrypoint has no HTTP session API. \\
WorldPlay (Hunyuan) & 10 chunks; 157 frames, 6.54 s & CLI takes the full pose and target length. Its 4-latent loop retains text KV, rebuilds visual KV from selected history, then decodes the accumulated latents. \\
WorldPlay (Wan) & 10 chunks; 157 frames, 9.81 s & Default: 4 chunks (about 61 frames, 3.81 s). \texttt{num\_frames=961} preallocates 241-latent noise, but \texttt{num\_chunk} sets generated chunks. Pose and decode remain inside one invocation. \\
LingBot-World (v1 Fast) & 10 chunks; 117 frames, 7.31 s & The CLI takes a complete camera trajectory and owns the 3-latent loop. It commits clean-history KV per chunk, then assembles latents for whole-sequence VAE decoding. \\
Matrix-Game 3.0 & 10 segments; 417 frames, 24.53 s & Normal mode constructs a full action/pose trajectory from \texttt{num\_iterations}; interactive mode reads one stdin action per segment but fixes the total count. Optional async VAE uses a separate GPU id. \\
SANA-WM & 40 chunks; 960 frames, 60.00 s & Native 24-frame callbacks use three CUDA streams for Stage-1, refiner, and decoder on one GPU. The 60-s run completed without OOM; the camera/action plan is supplied to the invocation. \\
DreamX-World & 84 chunks; 1,005 frames, 62.81 s & The AR entrypoint runs 3-latent chunks for a preset length and action sequence, then decodes the full sequence. The public run reached 81,037 MiB on an 80-GB A100. \\
AlayaWorld & 45 chunks; 1,440 frames, 60.00 s & Uses a precomputed camera path and full-sequence finalization. Retaining DiT, Gemma, and DA3 during decode reached about 80 GB on rank 0; releasing them reduced backend decode from 115.77 s to 33.08 s. \\
LingBot-World v2 & 10 chunks; 157 frames, 9.81 s attempted & The 4-latent CLI completed all 10 DiT chunks, then rank-0 whole-sequence VAE OOMed on an 854-MiB allocation. A 3-latent, 10-chunk run produced 117 frames (7.31 s). \\
ABot-World & 10 blocks; 117 frames, 9.75 s & The public call expands an action list before its block loop; rolling video KV and TAEHV cache persist across blocks. Ten blocks completed on one GPU, but the entrypoint lacks the WMI session API. \\
\bottomrule
\end{tabularx}
\end{table}

\noindent\textbf{Cache construction and device placement.}
For chunk-autoregressive backends, layerwise KV is initialized inside the model pipeline on the device that executes the DiT, then updated by a clean-context commit or rebuilt from the selected history at each native boundary. Matrix-Game 2.0 and ABot-World use a single visible CUDA device for visual/action/text KV and their causal-VAE feature cache. Hunyuan retains text KV and rebuilds visual KV from selected history on the active sequence-parallel ranks; WorldPlay Wan keeps transformer state on its sequence-parallel ranks and spatially shards VAE decoding across ranks before rank 0 assembles frames. LingBot v1/v2 keep per-layer self-KV and static cross-KV with the FSDP/Ulysses execution ranks, while their public CLI concatenates latents and performs the full VAE decode on rank 0. DreamX-World keeps its 30-layer self/cross caches on its single CUDA device and decodes the complete latent sequence there. SANA-WM's Stage-1 cache, refiner KV, and causal-decoder features share one GPU; its three CUDA streams overlap stages on that same device. Matrix-Game 3.0 has no cross-segment transformer KV: it retains the clean latent list, selects five field-of-view latents for each segment, and keeps the synchronized VAE state on rank 0; its optional asynchronous decoder runs on the extra device with id equal to the distributed world size. AlayaWorld carries its sink/history and spatial-memory state across context-parallel ranks, then performs final full-sequence VAE decoding on rank 0. These placements explain why the decoder and attention history can compete for memory even when the denoising loop itself advances one chunk at a time.

These observations point to two separate constraints. First, many public entrypoints keep the rollout loop inside a command-line or web invocation: WorldPlay (Hunyuan and Wan), LingBot v1/v2, DreamX-World, and AlayaWorld take a complete trajectory or length before running; Matrix-Game 3.0's interactive alternative blocks on stdin and still fixes the number of segments. Matrix-Game 2.0 and ABot-World already generate and decode native chunks, while SANA-WM exposes chunk callbacks. Thus, the adaptation problem is not the absence of chunked algorithms in every model; it is the lack of a common session boundary that lets the WorldProbe runner select a new action from each newly generated observation while preserving model state.

Second, some public paths retain large time-dependent tensors or overlap generation state with final decoding. In LingBot-World v2, the 4-latent, 10-chunk run completed its denoising loop but failed during the final rank-0 VAE decode on an 854-MiB allocation. In separate three-minute, 480p LingBot VAE measurements, GPU-side concatenation of decoded FP32 frames accounts for about 12.5 GiB; full-history self-KV in LingBot v1 also OOMed on 2- or 4-GPU runs. Our long-horizon configuration uses an 18-latent local window with 6 sink latents and streams decoded uint8 chunks to the artifact writer. In DreamX-World, the reproduced public VAE path reached 81,037 MiB; incremental chunk decoding reduced the corresponding driver peak to 49,795 MiB in the same 84-chunk comparison. AlayaWorld reached roughly 80 GB on rank 0 when generation models remained resident during finalization, so the driver follows the upstream release-before-decode boundary. These cases show why increasing a preset frame count alone does not provide a stable long-horizon interface: full-video buffers and unbounded history can consume the memory needed by later chunks or by the final decoder.

The implementation therefore preserves the upstream chunk computation and its cache semantics while bounding session state where the model supports it: rolling or field-of-view-selected KV replaces full-history attention when applicable, causal VAE state is carried across chunk boundaries, and decoded chunks are consumed incrementally by WMI. Each session still has a configured finite chunk bound; the design supports the paper's long-horizon rollouts without assuming an infinite GPU-resident sequence.

\subsection{Target Tracking, Geometry, and Rollout Screening}
\label{app:tracking-geometry}

At initialization, SAM3.1 receives the text description of the target and establishes its mask and tracker memory. After each generated chunk, the updated observation is passed through the tracker to recover the target location. A geometric gate rejects masks whose frame-to-frame displacement is implausibly large, reducing abrupt target switches during long camera motion. The agent receives the resulting target direction together with recent frames and task instructions.

DA3 estimates depth for the current observation, from which WorldProbe builds a coarse local voxel representation. The geometry module measures free space along the current forward direction; it does not maintain a global reconstruction. If the estimated clearance is no greater than the forward displacement in the preceding chunk, the controller overrides a forward action with a turn toward the tracked target. This near-field gate handles immediate obstacle clearance, while the task agent remains responsible for global route choice.

The runner records the requested and executed actions, tracker status, termination reason, and artifact metadata for each rollout. A generated video is eligible for scoring only after it passes the minimum-duration rule and coarse screening for severe visual or trajectory anomalies. The local visual and content measures, ViPE-based pose selection, long-range memory comparisons, and VLM evolution scores are then computed from the finalized rollout as described in Section~\ref{sec:metrics}.

\clearpage
\section{Representative Rollout Trajectories with Memory and Evolution Diagnostics}
\label{app:case-trajectories-memory-evolution}

\begin{figure}[p]
\centering
\includegraphics[width=\linewidth]{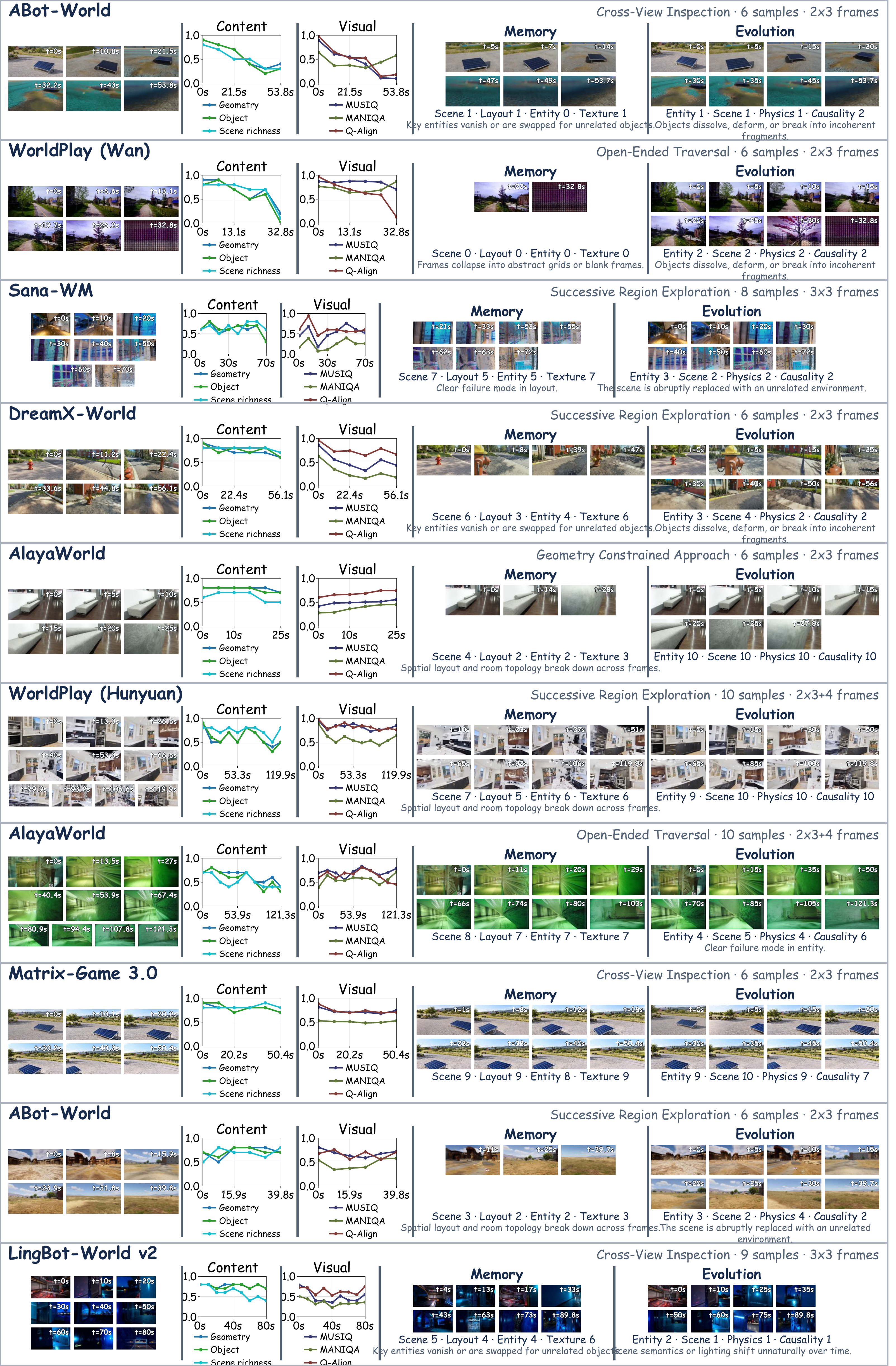}
\caption{\textbf{Representative rollout trajectories with local, memory, and evolution diagnostics.} Each row shows sampled rollout frames (left), Content and Visual metric curves, and Memory and Evolution comparison frames with dimension scores and failure explanations. Gray headers report the probing task, sample count, and frame layout.}
\label{fig:case-trajectories-memory-evolution}
\end{figure}

Fig.~\ref{fig:case-trajectories-memory-evolution} extends the representative local-quality cases in Fig.~\ref{fig:local-cases} with per-rollout Content and Visual metric curves, together with Memory and Evolution diagnostic frames and VLM rationales for ten models spanning diverse tasks and failure modes.

\end{document}